\documentclass[runningheads]{llncs}

\usepackage{eccv}

\usepackage{eccvabbrv}

\usepackage{graphicx}
\usepackage{booktabs}
\usepackage[accsupp]{axessibility}  % Improves PDF readability for those with disabilities.

\usepackage{amsmath,amsfonts,bm}

\def\eqref#1{equation~\ref{#1}}
\def\1{\bm{1}}

\def\vc{{\bm{c}}}
\def\vd{{\bm{d}}}

\def\vh{{\bm{h}}}

\def\vp{{\bm{p}}}

\def\vr{{\bm{r}}}

\def\vx{{\bm{x}}}

\def\mA{{\bm{A}}}

\def\mC{{\bm{C}}}

\def\mE{{\bm{E}}}
\def\mF{{\bm{F}}}

\def\mH{{\bm{H}}}
\def\mI{{\bm{I}}}

\def\mM{{\bm{M}}}

\def\mP{{\bm{P}}}

\def\mS{{\bm{S}}}

\def\mW{{\bm{W}}}
\def\mX{{\bm{X}}}

\DeclareMathAlphabet{\mathsfit}{\encodingdefault}{\sfdefault}{m}{sl}
\SetMathAlphabet{\mathsfit}{bold}{\encodingdefault}{\sfdefault}{bx}{n}

\newcommand{\R}{\mathbb{R}}

\usepackage{comment}
\usepackage{amsmath,amssymb,amsfonts,mathabx,mathbbol}
\usepackage{acronym}
\usepackage{enumitem}
\PassOptionsToPackage{dvipsnames,table}{xcolor}
\usepackage{url}
\usepackage{balance}
\usepackage{xspace}
\usepackage{setspace}
\usepackage[skip=3pt,font=small]{subcaption}
\usepackage[skip=3pt,font=small]{caption}
\usepackage{caption}
\usepackage[misc]{ifsym}
\usepackage{tabularx}
\usepackage{multirow,multirow,array,makecell}
\usepackage{overpic}
\usepackage{wrapfig}
\usepackage{pifont}
\usepackage{url}
\usepackage{tabularx,colortbl}         % professional-quality tables
\usepackage[utf8]{inputenc}                     % allow utf-8 input
\usepackage[T1]{fontenc}                        % use 8-bit T1 fonts
\usepackage{nicefrac}                           % compact symbols for 1/2, etc.
\usepackage[nopatch=footnote]{microtype}        % microtypography
\usepackage[noend,ruled,vlined]{algorithm2e}
\usepackage{algorithmic}
\usepackage{rotating}
\usepackage{adjustbox}
\newcommand{\sota}{\text{state-of-the-art}\xspace}
\newcommand\supp{\textit{supplementary}\xspace}

\makeatletter
\renewcommand\paragraph{\@startsection{paragraph}{4}{\z@}%
                       {-6\p@ \@plus -4\p@ \@minus -4\p@}%
                       {-0.5em \@plus -0.22em \@minus -0.1em}%
                       {\normalfont\normalsize\bfseries}}
\makeatother

\graphicspath{{figures/}}

\acrodef{nlp}[NLP]{Natural Language Processing}
\acrodef{vos}[VOS]{Video Object Segmentation}

\acrodef{gru}[GRU]{Gated Recurrent Unit}
\acrodef{mhsa}[MHSA]{Multi-Head Self-Attention}
\acrodef{mha}[MHA]{Multi-Head Attention}
\newcommand{\sa}{\text{Slot-Attention}\xspace}

\acrodef{nerf}[NeRF]{Neural Radiance Field}
\newcommand{\model}{\textsc{SlotLifter}\xspace}

\acrodef{ari}[ARI]{Adjusted Rand Index}
\acrodef{msc}[MSC]{Mean Segmentation Covering}
\acrodef{miou}[mIoU]{mean Intersection over Union}
\acrodef{mse}[MSE]{Mean Squared Error}

\definecolor{scope}{RGB}{103,78,167}
\definecolor{semantic}{RGB}{230,145,56}
\definecolor{type}{RGB}{153,0,0}
\definecolor{uclablue}{rgb}{0.15, 0.45, 0.68}
\definecolor{cvprblue}{rgb}{0.21,0.49,0.74}
\definecolor{revision}{RGB}{0,0,0}

\makeatletter
\renewcommand*{\@fnsymbol}[1]{\ensuremath{\ifcase#1\or *\or \dagger\or \ddagger\or
   \mathsection\or \mathparagraph\or \|\or **\or \dagger\dagger
   \or \ddagger\ddagger \else\@ctrerr\fi}}
\makeatother

\usepackage{hyperref}

\usepackage{orcidlink}

\begin{document}

\title{\model: Slot-guided Feature Lifting for Learning Object-centric Radiance Fields} 

\titlerunning{Slot-guided Feature Lifting for Learning Object-centric Radiance Fields}

\author{Yu Liu$^{\star\dagger}$\inst{1,2}\orcidlink{0009-0002-2367-211X} \and
Baoxiong Jia$^\star$\inst{1}\orcidlink{0000-0002-4968-3290} \and
Yixin Chen\inst{1}\orcidlink{0000-0002-8176-0241} \and
Siyuan Huang\inst{1}\orcidlink{0000-0003-1524-7148}
}
\authorrunning{Y. Liu and B. Jia et al.}

\institute{
State Key Laboratory of General Artificial Intelligence, BIGAI \and
Department of Automation, Tsinghua University\\
$^\star$Equal contribution $^\dagger$Work done as an intern at BIGAI 
\url{https://slotlifter.github.io}}

\maketitle

% \blfootnote{$^\star$Equal contribution. $^\dagger$Work done during internship at BIGAI.}

\begin{abstract}
The ability to distill object-centric abstractions from intricate visual scenes underpins human-level generalization. Despite the significant progress in object-centric learning methods, learning object-centric representations in the 3D physical world remains a crucial challenge. In this work, we propose \model, a novel object-centric radiance model addressing scene reconstruction and decomposition jointly via slot-guided feature lifting. Such a design unites object-centric learning representations and image-based rendering methods, offering \sota performance in scene decomposition and novel-view synthesis on four challenging synthetic and four complex real-world datasets, outperforming existing 3D object-centric learning methods by a large margin. 
Through extensive ablative studies, we showcase the efficacy of designs in \model, revealing key insights for potential future directions.
\end{abstract}
% Specifically, our method achieves a significant improvement of \textbf{4+ PSNR} and \textbf{20+ ARI} over previous SOTA methods, thus showcasing its effectiveness and efficiency. 

\section{Introduction}\label{sec:introduction}
The sense of objectness has been crucial to human cognition and generalization capabilities~\cite{spelke2007core,lake2017building}. Despite recent advances in visual perception~\cite{he2016deep,dosovitskiy2020image,caron2021emerging,oquab2023dinov2}, achieving this generalization capability remains an unsolved challenge for existing models~\cite{greff2020binding}. The pivotal role of object-centric understanding in human cognition necessitates models that can extract symbol-like object abstractions from complex visual signals, forming object-centric representations without supervision.

Recent years have witnessed substantial progress in object-centric learning~\cite{greff2019iodine, locatello2020sa, singh2021slate, jia2022improving}. These methods aim to disentangle visual scenes into object-like entities for object-oriented reasoning and manipulation. Despite the remarkable progress made, existing approaches predominantly focus on 2D images. Since 2D images provide only partial views of the 3D physical world, object representations learned in the 2D domain are easily bound to 2D object attributes like colors~\cite{jia2022improving}, neglecting crucial information about object shape, geometry, and spatial relationships. Given the importance of these 3D attributes in representing the physical world, it is essential for models to form object abstractions in 3D environments to enhance understanding and interaction with the real world~\cite{sajjadi2022object,driess2023palm}.

To fulfill this goal, various attempts have been made to combine object-centric methods such as \sa~\cite{locatello2020sa} with 3D representations. Among them, multi-view image representations of 3D scenes~\cite{yu2021unsupervised,stelzner2021decomposing,smith2023colf} show competitive results on synthetic datasets given their effectiveness in preserving detailed object information. Nonetheless, translating the success of these methods from synthetic data to real-world scenarios has been proven to be non-trivial~\cite{seitzer2023bridging}. Specifically, aggregating information from multi-view real images and drawing correspondences between them naturally requires more intricate model designs. Meanwhile, decoding from object-centric representations to 3D (\eg, novel views) places higher demands on the learned representations (\ie, slots) as it now needs to infer about the 3D scene from a series of calibrated partial view projections.
Recently, OSRT~\cite{sajjadi2022object} scales up the dimensions of slots and reconstructs scenes with a Transformer-based encoder-decoder architecture, demonstrating powerful decomposition and reconstruction ability in complex 3D scenes. However, its success is built at the cost of inadmissible data and computation demands (64 TPUv2 chips for 7 days on 1M scenes). This urges the need for methods to effectively align information from calibrated multi-view images and reconstruct 3D scenes from the compressed object-centric representations.

In this work, we present \model, a novel approach to learning object-centric representations in 3D scenes, inspired by recent advances in image-based rendering methods~\cite{yu2021pixelnerf, wang2021ibrnet, chen2021mvsnerf, gao2023surfelnerf, zhang2022nerfusion, varma2022attention, cong2023enhancing, xu2022point}. In contrast to previous object-centric methods that focus solely on decoding information from slots, our method leverages lifted 2D input-view feature(s) to initialize 3D point features, which interact with the learned slot representations via a cross-attention-based transformer for predicting volume rendering parameters. This design enhances the granularity of details for novel-view synthesis while providing more explicit guidance for slot learning. Additionally, with no auxiliary losses needed, \model relies only on the reconstruction loss and naturally requires less sampling overheads during training compared with existing 3D object-centric learning models like uORF and OSRT. This results in significantly fewer computational resources needed ($\sim$5x faster) to achieve desirable outcomes. Through comprehensive experiments on four challenging synthetic and four complex real-world datasets, we observe consistent and significant performance improvement of \model over existing 3D object-centric models on both scene decomposition ($\sim$10+ ARI) and novel-view synthesis ($\sim$2+ PSNR). We further show the effectiveness of each module through extensive ablative analyses and discussions, offering new insights into developing object-centric learning techniques for complex 3D scenes. In summary, our main contributions are as follows:
\begin{enumerate}[nolistsep, noitemsep]

\item We propose \model, a novel model for unsupervised object-centric learning in 3D scenes that effectively aggregates multi-view features for object-centric decoding via an innovative slot-guided feature lifting design.

\item We comprehensively evaluate \model across four challenging synthetic and four real-world benchmarks. Our results consistently show that \model significantly outperforms existing methods in both scene decomposition and novel-view synthesis, achieving \sota performance.

\item We conduct extensive ablative analyses demonstrating \model's potential in object-centric learning and image-based rendering, especially given its superior performance on established complex real-world datasets (\eg, ScanNet and DTU) against \sota image-based rendering methods. We anticipate that our findings will stimulate further advancements in overcoming current limitations of 3D object-centric models.
\end{enumerate}

\section{Related Work}\label{sec:related_work}
\paragraph{Object-centric Learning}
Prior studies in object-centric learning~\cite{greff2016tagger,eslami2016attend,greff2017neural,greff2019iodine,burgess2019monet,crawford2019spatially,engelcke2019genesis,lin2020space,bear2020learning,locatello2020sa, zoran2021parts} have demonstrated proficiency in disentangling visual scenes into object-centric representations primarily on synthetic datasets, but they often struggle with handling complex real-world scenes. Notably, \sa~\cite{locatello2020sa} has fostered many powerful variants~\cite{singh2021slate, lamb2021transformers, choudhury2021unsupervised, caron2021emerging, wang2022self, du2021unsupervised, henaff2022object,seitzer2023bridging,jia2022improving,wang2023object} across various tasks and domains. However, these methods typically focus solely on learning object-centric representations from static images, thereby overlooking motion and 3D geometry information crucial for decomposing real-world complex scenes in an object-centric manner. 
Recognizing the potential benefits of motion information,~\cite{kipf2021conditional,elsayed2022savi++,singh2022simple} utilize video data to carve out object representations, demonstrating the effectiveness of the additional information provided beyond static images in the context of object-centric learning. Nonetheless, the use of 3D geometry information for object-centric learning has been largely left untouched. In this work, we pinpoint these crucial aspects by integrating advancements in image-based rendering with \sa, aiming to improve the acquisition of 3D object-centric representations within complex real-world environments.

\paragraph{Novel-view Synthesis with NeRFs}
Recent advances in \ac{nerf} methods~\cite{mildenhall2021nerf, barron2022mip, tancik2022block} have shown notable success in novel-view synthesis and 3D scene reconstruction. However, a significant drawback of these methods is the scene-specific long training time needed for optimizing each scene. 
% Various approaches~\cite{chen2022tensorf, muller2022instant, fridovich2022plenoxels} have been introduced to reduce this training overhead but these approaches still focus on per-scene optimization. 
The demand for better time efficiency has led to the emergence of generalizable NeRF methods~\cite{yu2021pixelnerf, wang2021ibrnet, chen2021mvsnerf, gao2023surfelnerf, zhang2022nerfusion, varma2022attention, cong2023enhancing, xu2022point,chen2024single}. These methods aim to synthesize novel views based on given images of scenes without per-scene optimization. For instance, PixelNeRF~\cite{yu2021pixelnerf} and IBRNet~\cite{wang2021ibrnet} adopt volume rendering techniques, using features from nearby views to reconstruct novel views. MVSNeRF~\cite{chen2021mvsnerf} constructs cost-volumes from nearby views for novel-view rendering. PointNeRF~\cite{xu2022point} leverages latent point clouds as anchors for radiance fields to improve both efficiency and performance. GNT~\cite{varma2022attention} uses a transformer to integrate features from different views and demonstrates the powerful capability for generalizable novel-view synthesis. In contrast to these methods, \model leverages an object-centric multi-view feature aggregation module and point-slot mapping module to more effectively encode 3D complex scenes for generalizable novel-view synthesis.

\paragraph{3D Object-centric methods}
Previous methods~\cite{yu2021unsupervised, stelzner2021decomposing, sajjadi2022object, smith2023colf} have attempted to extend \sa to 3D scenes for scene decomposition and novel-view synthesis. uORF~\cite{yu2021unsupervised}, ObSuRF~\cite{stelzner2021decomposing}, sVORF~\cite{qi2023slot}, and uOCF~\cite{luo2024unsupervised} combine \sa (or its variants) with NeRF~\cite{mildenhall2021nerf} and use rendering losses as objectives for unsupervised slot learning. Additionally, OSRT~\cite{sajjadi2022object} and COLF~\cite{smith2023colf} further introduce \sa into the light field model to improve both model performance and inference speed. Nevertheless, uORF, COLF, and uOCF necessitate extra auxiliary losses, such as adversarial loss and LPIPS loss with a prolonged training period, which prevents downsampling rays and needs more computation. ObSuRF and uOCF require training with depth as a guidance signal. OSRT suffers from the heavy computation and training overhead required for properly reconstructing views from input pose and image embeddings. In contrast, \model lifts the 2D multi-view feature to 3D and uses these point features to query multi-view information from the learned slots effectively. From our experiments, \model not only outperforms previous 3D object-centric methods for unsupervised scene decomposition and novel-view synthesis but also obtains higher training efficiency.

\section{\model}\label{sec:method}
\begin{figure*}[t!]
    \centering
    % \fbox{\rule[0cm]{0cm}{7cm}\rule[0cm]{\linewidth}{0cm}}
    \resizebox{\linewidth}{!}{\includegraphics[width=\linewidth]{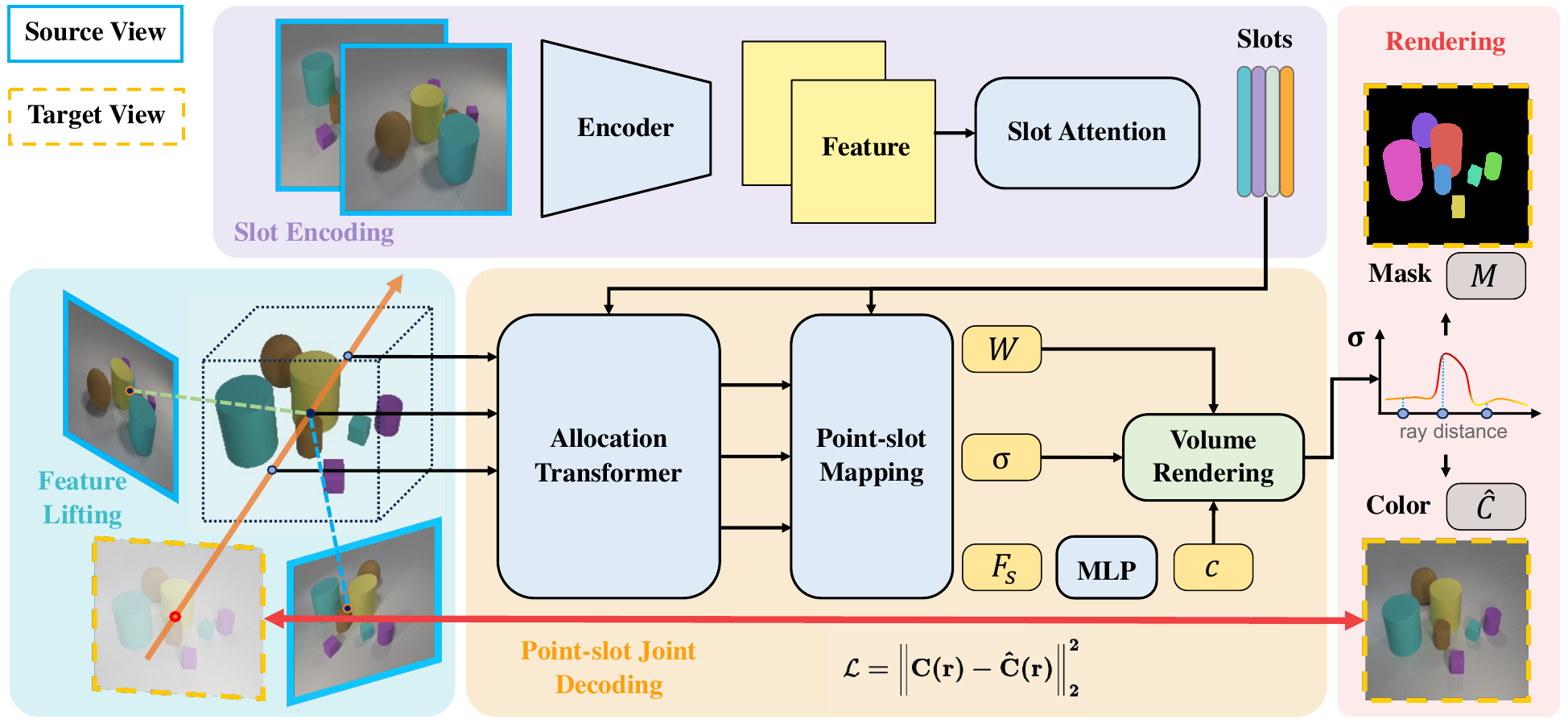}}
    \caption{\textbf{\model overview.} \model extracts slots from input view(s) during slot encoding. It then lifts 2D feature maps of input view(s) to initialize 3D point features, which serve as queries in the allocation transformer for point-slot joint decoding. This process yields the point-slot mapping $\mW_p$, density $\sigma$, and the slot-aggregated point feature $\mF_s$ via an attention layer. Finally, \model uses these results for rendering novel-view images and segmentation masks via volume rendering.}
    \label{fig:model_arch}
    \vspace{-.15in}
\end{figure*}

In this section, we introduce our model, \model, that combines object-centric learning modules with image-based rendering techniques. Our goal is to effectively learn scene reconstruction and decomposition by reconstructing input-view image(s).
We present an overview of our \model model in~\cref{fig:model_arch}.
% Our method is based on \sa, incorporating the feature lifting operation used in generalizable NeRFs for image-based rendering. Specifically, we initialize 3D point features by projecting 3D points onto 2D feature maps,  then decoding slots with lifted 3D point features jointly. 
\subsection{Background}\label{sec:method:background}
\paragraph{Object-centric learning via \sa} Given $N$ input feature vectors $\mX\in\R^{N\times D_f}$, 
\sa~\cite{locatello2020sa} maps them to a set of $K$ output vectors (\ie, slots) $\mS\in\R^{K\times D_s}$ via an iterative attention mechanism. The $K$ slots compete to explain the input features $\mX$ by computing the attention matrix $\mA$ between $\mS$ and $\mX$. The attention matrix is then used to aggregate feature vectors $\mX$ using a weighted mean. These aggregated features are embedded into slots $\mS$ by iteratively updating as follows:
\begin{equation}
\begin{aligned}
& \mA = \mathrm{softmax}\left(\frac{k(\mX)\cdot q(\tilde{\mS})^T}{\sqrt{D}}\right) \\
\mS = \mathcal{U}_\theta(\tilde{\mS}&,\ \mW^T v(\mX)),\ \mathrm{where}\ \mW_{i,j}=\frac{\mA_{i,j}}{\sum_{m=1}^{N}\mA_{m,j}}.
\end{aligned}
\label{equ:sa}
\end{equation}
$q(\cdot),\ k(\cdot),\ v(\cdot)$ are linear projections, $\tilde{\mS}$ denotes random initialized slots and $\mathcal{U}_\theta(\cdot)$ represents the iterative update function often implemented with GRU~\cite{cho2014learning}, LayerNorm~\cite{ba2016layer} and a residual MLP. As pointed out by Jia \etal~\cite{jia2022improving}, this iterative update process could be susceptible to instability when propagating gradients back into the iterative process. They therefore proposed a bi-level method, dubbed BO-QSA, to improve the optimization within \sa with learnable slot initialization instead of random sampled ones. 

\paragraph{Neural Radiance Fields} Given rays $\{\vr\}$ of a camera view, \ac{nerf} samples points along each ray and represent 3D scenes with a feature field $\mF_\Theta:(\vx,\vd)\rightarrow(\vc,\sigma)$ mapping the 3D location $\vx$ and the view direction $\vd$ to color $\vc$ and volume density $\sigma$, and then renders the color of each ray via volume rendering~\cite{max1995optical}: 
\begin{equation}\label{equ:render}
\hat{\mC}(\vr) = \sum_{i=1}^{N}T_i[1-\exp(-\sigma_i\delta_i)]\vc_i,
\end{equation}
where $T_i = \exp(-\sum_{j=1}^{i-1}\sigma_j\delta_j)$ and $\delta_i$ is the distance between adjacent volumes along a ray. While \ac{nerf} achieves impressive novel-view synthesis quality, it adds stringent demands on model training given the number of points needed for approximating $\hat{\mC}(\vr)$ in~\cref{equ:render}. It also exhibits no generalization capabilities as each scene is optimized individually without shared prior knowledge.

\subsection{Slot-guided Feature Lifting}\label{sec:method:slot_lifting}
\subsubsection{Scene Encoding} To render a novel target view $\mI_t$, we leverage \sa to encode scene representations from $L$ source view(s) $\{\mI_l\}_{l=1}^L$ ($L=1$ for single-view input) and lift 2D features to 3D for approximating the latent feature field $\mF_{\Theta}$. We start by extracting 2D feature maps $\{\mF_l^{\mathrm{2D}} \in \mathbb{R}^{H\times W\times D_{\text{f}}}\}_{l=1}^L$ from each source view. Next, we follow~\cref{equ:sa} to obtain object-centric scene features $\mS = \{\bm{s}_1,\cdots,\bm{s}_{K}\}\in\mathbb{R}^{K\times D_{\text{s}}}$ from these input 2D features via \sa. Inspired by image-based rendering methods, we consider constructing an additional 3D scene feature field by lifting 2D input-view features for capturing the fine-grained details in the input. Specifically, for target view $\mI_t$, we sample points $\mP\in \mathbb{R}^{N\times 3}$ along each ray $\vr$ and project each 3D point $\vp=(x,y,z)$ onto the image coordinates $\pi(\vp)=(x',y')$ to obtain its set of corresponding 2D features $\mF_{\mathrm{lift}}(\vp)$ by:
\begin{equation*}
\mF_\mathrm{lift}(\vp)=\left[\mF^{\mathrm{2D}}_1[\pi(\vp)],\cdots,\mF^{\mathrm{2D}}_L[\pi(\vp)]\right].
\end{equation*}
Without adding further ambiguity to the notations, we use $\mF_{\mathrm{lift}}\in\mathbb{R}^{N\times L\times D_{\text{f}}}$ to represent the feature field obtained for all points in $\mP$. After obtaining the lifted point features $\mF_{\mathrm{lift}}$, we pool the multi-view features to obtain 3D point features:
\begin{equation}
\label{eq:fusion}
\mF_p = \mathrm{MLP}([\mathrm{Mean}(\mF_\mathrm{lift}), \mathrm{Var}(\mF_\mathrm{lift})]) + \mE_p,
\end{equation}
where $\mE_p\in R^{N\times D_p}$ are positional embeddings for preserving the spatial information of 3D points. Notably, for single-view input, we ignore the variance term and let $\mF_p = \mathrm{MLP}(\mF_{\mathrm{lift}}) + \mE_p$. This feature serves a similar role as $\mF_{\Theta}$ discussed in~\cref{equ:render}, providing fine-grained 3D features with spatial location considered.

\subsubsection{Point-slot Mappping} After scene encoding, given the slots $\mS$ and the point features $\mF_p$, we design a point-slot joint decoding process to leverage both point and slot features for rendering. First, we calculate the point-slot mapping $W_p$, identifying the points that a slot $\bm{s}_i \in \mS$ contributes to. Specifically, we use a cross-attention-based allocation transformer, leveraging point features $\mF_p$ as queries and slot representations $\mS$ as keys and values to allocate slots to 3D points. As some points map to vacant areas in the 3D space, we add an additional learnable empty slot $\bm{s}_{\emptyset}$ for these vacant points to query from. This process could be summarized as:
% We use a 1D convolution and a self-attention layer over point features after every cross-attention layer to further enhance information extraction from slots to 3D points:
\begin{align*}
\mS' = \{\bm{s_\emptyset}, \bm{s}_1, \cdots, \bm{s}_k\},\ \tilde\mF_p = \mathrm{CrossAttn}(Q=\mF_p,KV=\mS').
\end{align*}
After this process, the 3D point features $\tilde{\mF}_{p}$  contain information queried from object-centric slot representations. Finally, we obtain the point-slot mapping and the slot-aggregated point feature $\mF_{s}$  via an attention layer following:
\begin{equation*}
    \begin{aligned}
    \mF_s = \mW_p\mS',\ \text{where}\ \mW_p = \mathrm{softmax}\left(\frac{q(\tilde\mF_p)\cdot k(\mS')^T}{\sqrt{D}}\right).
    % \mA_p = q(\tilde\mF_p)\cdot k(\mS')^T,\ \mW = \mathrm{softmax}(\mA_p),\ \mF_s = \mW\mS'
    \end{aligned}
\end{equation*}
We use $q(\cdot),\ k(\cdot)$ to denote linear projections, $\mW_p\in\mathbb{R}^{N\times (K+1)}$ for the mapping weights from slots to points, $D$ for the latent feature dimension. In essence, this process aims to obtain decodable 3D representations from learned slots. We can find the corresponding slot mapping (\ie, contribution) weight from $\mW_{p}^i$ for each 3D point $\vp_i$, thereby predicting its slot assignment for scene decomposition.

\subsubsection{Slot-based Density} For notation purposes, we use $\mA_p=q(\tilde\mF_p)\cdot k(\mS')^T$ throughout the subsequent texts for simplicity. To provide more direct guidance to slots, we use the attention weights $\mA_p$ from the mapping module to estimate the density value following~\cite{yu2021unsupervised}:
\begin{equation}\label{equ:density}
    \sigma_i = \mathrm{sum}(\mW_p^{i,1:K+1} \odot \mathrm{ReLU}(\mA_p^{i, 1:K+1})),
\end{equation}
where $i$ denotes $i$-th point, $\odot$ denotes Hadamard production, and $\mA_p^{i, 1:K+1}$ denotes the attention weights of the last K slots, ignoring the first empty slot in $\mS'$.  We add a ReLU layer over $\mA_{p}$ to suppress the contribution of slots less related to a specific point $\mF_p^i$ in
density prediction.
Finally, we add $\mF_{s}$ with the positional embedding $\mE_{p}$ and pass it into an MLP for predicting colors ${\vc}$. Similarly, given the 3D point-slot mapping weight $\mW_p^i\in R^{K}$ of each point, \model is able to render 2D segmentation masks $\mM$ using the same rendering scheme:
\begin{equation}
    \label{equ:rendering}
    \begin{aligned}
    \vc = \mathrm{MLP}(\mF_s + \mE&_p),\ \mC(\vr) = \sum_{i=1}^{N}T_i[1 - \exp(-\sigma_i\delta_i)]\vc_{i},\\
    \mM(\vr) &= \sum_{i=1}^{N}T_i[1-\exp(-\sigma_i\delta_i)]\mW_p^i,
    \end{aligned}
\end{equation}
where $T_i = \exp(-\sum_{j=1}^{i-1}\sigma_j\delta_j)$ and $\delta_i$ is the distance between adjacent volumes along a ray following~\cref{equ:render}.

\subsection{Training}\label{sec:method:training}
\paragraph{Objective} For training, we utilize the mean squared error (MSE) between the rendered rays $\mC(r)$ and the ground truth colors $\hat{\mC}(r)$ as our learning objective: 
$$
\mathcal{L}_{\text{recon}}=\Vert \mC(r)-\hat{\mC}(r)\Vert^2.
$$
\paragraph{Random Masking} Although incorporating feature lifting into 3D object-centric learning improves the utilization of 3D information, it also poses a significant problem. Since both lifted point features $\mF_p$ and slot features $\mS$ originate from 2D multi-view images, the model can converge to degenerate scenarios, relying solely on lifted features for rendering and ignoring the information in slots. We avoid this degenerate case by randomly masking the lifted features in the sampled points, using only positional embeddings $\mE_p$ for these points to enforce alignment between slots and 3D point grids. In implementation, we use a cosine annealing schedule on the masking ratio from 0.99 to 0 for 30K steps.

\section{Experiment}\label{sec:exp}
We present experimental results of \model on \textbf{4 synthetic} and \textbf{4 complex real-world} datasets, evaluating its capability in novel view synthesis and unsupervised scene decomposition. The experimental settings are as follows:

\paragraph{Datasets} For synthetic scenes, we evaluate \model on 3 commonly used datasets CLEVR-567, Room-Chair, and Room-Diverse proposed by uORF~\cite{yu2021unsupervised}. We further select a more complex variant of Room-Diverse, Room-Texture~\cite{luo2024unsupervised}, that provides synthetic rooms with real objects from ABO~\cite{collins2022abo} for evaluating 3D object-centric learning.
For complex real-world scenes, we use Kitchen-Shiny~\cite{luo2024unsupervised}, Kitchen-Matte~\cite{luo2024unsupervised}, ScanNet~\cite{dai2017scannet}, and DTU MVS~\cite{jensen2014large} to evaluate models' capability on novel-view synthesis and scene decomposition. 

\paragraph{Metrics} We evaluate the quality of novel-view synthesis with three common metrics: LPIPS~\cite{zhang2018unreasonable}, SSIM~\cite{wang2004image}, and PSNR. In particular, we use $\mathrm{LPIPS}_{\text{alex}}$ for synthetic scenes and $\mathrm{LPIPS}_{\text{vgg}}$ for real-world scenes to be consistent with previous methods. Following~\cite{yu2021unsupervised, smith2023colf}, we evaluate the quality of scene decomposition with four metrics: \ac{ari}, FG-ARI (\ie, ARI computed only on foreground objects), NV-ARI (\ie, ARI on novel views), and NV-FG-ARI. 

\subsection{Object-centric Learning in Synthetic Scenes}\label{sec:exp:synthetic}
\paragraph{Setup} To perform a fair comparison between \model and existing methods, we follow the setup of uORF~\cite{yu2021unsupervised} and use only \textbf{one source view} as input to render the other novel views. As we only use a single source view, we modify the multi-view feature aggregation to $\mF_p = \mathrm{MLP}(\mF_\mathrm{lift}) + \mE_p$ as discussed in~\cref{sec:method:slot_lifting}. We train our model using the Lion~\cite{chen2023symbolic} optimizer with a learning rate of 5$\times10^{-5}$ for 250k iterations. We use a batch size of 4 and sample 1024 rays for each scene. 

\begin{table*}[t!]
    \caption{\textbf{Quantitative comparison for segmentation in synthetic scenes.} \model achieves the best performance on most metrics. Especially, when the dataset complexity increases (\eg, from Room-Chair to Room-Diverse), \model makes remarkable improvements (10+ ARI). We report all models with (mean $\pm$ standard deviation) across 3 experiment trials except for sVORF where we report the best performance ($\dagger$) adapted from the paper.} 
    \vspace{-8pt}
    \centering
    \label{tab:synthetic_seg}
    \resizebox{\linewidth}{!}{
    \begin{tabular}{cccccccccc}
    \toprule
    \multirow{3}[3]{*}{Method} & \multicolumn{3}{c}{CLEVR-567} & \multicolumn{3}{c}{Room-Chair}& \multicolumn{3}{c}{Room-Diverse}\\
    \cmidrule(lr){2-4}\cmidrule(lr){5-7}\cmidrule(lr){8-10}
    &3D metric &\multicolumn{2}{c}{2D metric} &3D metric &\multicolumn{2}{c}{2D metric} &3D metric &\multicolumn{2}{c}{2D metric} \\
    \cmidrule(lr){2-2}\cmidrule(lr){3-4}\cmidrule(lr){5-5}\cmidrule(lr){6-7}\cmidrule(lr){8-8}\cmidrule(lr){9-10}
    & NV-ARI$\uparrow$ & ARI$\uparrow$ & FG-ARI$\uparrow$ & NV-ARI$\uparrow$ & ARI$\uparrow$ & FG-ARI$\uparrow$ & NV-ARI$\uparrow$ & ARI$\uparrow$ & FG-ARI$\uparrow$ \\
    \midrule
    \sa~\cite{locatello2020sa} & - & 3.5$\pm$0.7 & \textbf{93.2$\pm$1.5} & - & 38.4$\pm$18.4 & 40.2$\pm$4.5 & - & 17.4$\pm$11.3 & 43.8$\pm$11.7 \\
    uORF~\cite{yu2021unsupervised} & {83.8$\pm$0.3} & 86.3$\pm$0.1 & 87.4$\pm$0.8 & 74.3$\pm$1.9 & 78.8$\pm$2.6 & 88.8$\pm$2.7 & 56.9$\pm$0.2 & 65.6$\pm$1.0 & 67.8$\pm$1.7 \\
    BO-uORF~\cite{jia2022improving} & 78.4$\pm$0.7 & {87.4$\pm$0.5} & 89.2$\pm$0.3 & {80.9$\pm$0.2} & 82.2$\pm$1.0 & {91.6$\pm$2.3} & {62.5$\pm$0.5} & {72.6$\pm$0.2} & {76.8$\pm$0.2} \\
    COLF~\cite{smith2023colf} & 55.8$\pm$0.1 & 69.0$\pm$0.4 & {92.4$\pm$1.7} & 80.7$\pm$0.1 & {85.6$\pm$0.04} & 89.8$\pm$0.1 & 52.5$\pm$0.3 & 66.5$\pm$0.4 & 64.7$\pm$0.7 \\
    \midrule
    \model & \textbf{87.0$\pm$2.5} & \textbf{93.7$\pm$1.1} & 91.3$\pm$1.6 & \textbf{89.7$\pm$0.5} & \textbf{92.6$\pm$0.3} & \textbf{91.9 $\pm$0.3} & \textbf{77.5$\pm$0.7} & \textbf{90.0$\pm$0.8} & \textbf{84.3$\pm$2.7} \\
    \midrule
    sVORF$^\dagger$ ~\cite{qi2023slot} &81.5 &82.7 &92.0 &87.0 &87.8 &\textbf{92.4} &75.6 &78.4 &86.6\\
    \textsc{SlotLifter}$^\dagger$ & \textbf{89.0} & \textbf{94.6} & \textbf{93.1} & \textbf{90.3} & \textbf{92.9} & 92.1 & \textbf{78.1} & \textbf{90.6} & \textbf{86.7} \\
    \bottomrule
    \end{tabular}
    }
    \vspace{-.15in}
\end{table*}
\paragraph{Baselines} We compare \model with previous state-of-the-art 3D object-centric methods including uORF~\cite{yu2021unsupervised}, COLF~\cite{smith2023colf}, and sVORF~\cite{qi2023slot}. We also report the results of the improved uORF (BO-uORF) introduced by Jia \etal~\cite{jia2022improving} as a competitive baseline in evaluating the results on these datasets. 

\begin{figure*}[t!]
    \centering
    % \fbox{\rule[0cm]{0cm}{12cm}\rule[0cm]{\linewidth}{0cm}}
    \resizebox{\linewidth}{!}{\includegraphics[width=\linewidth]{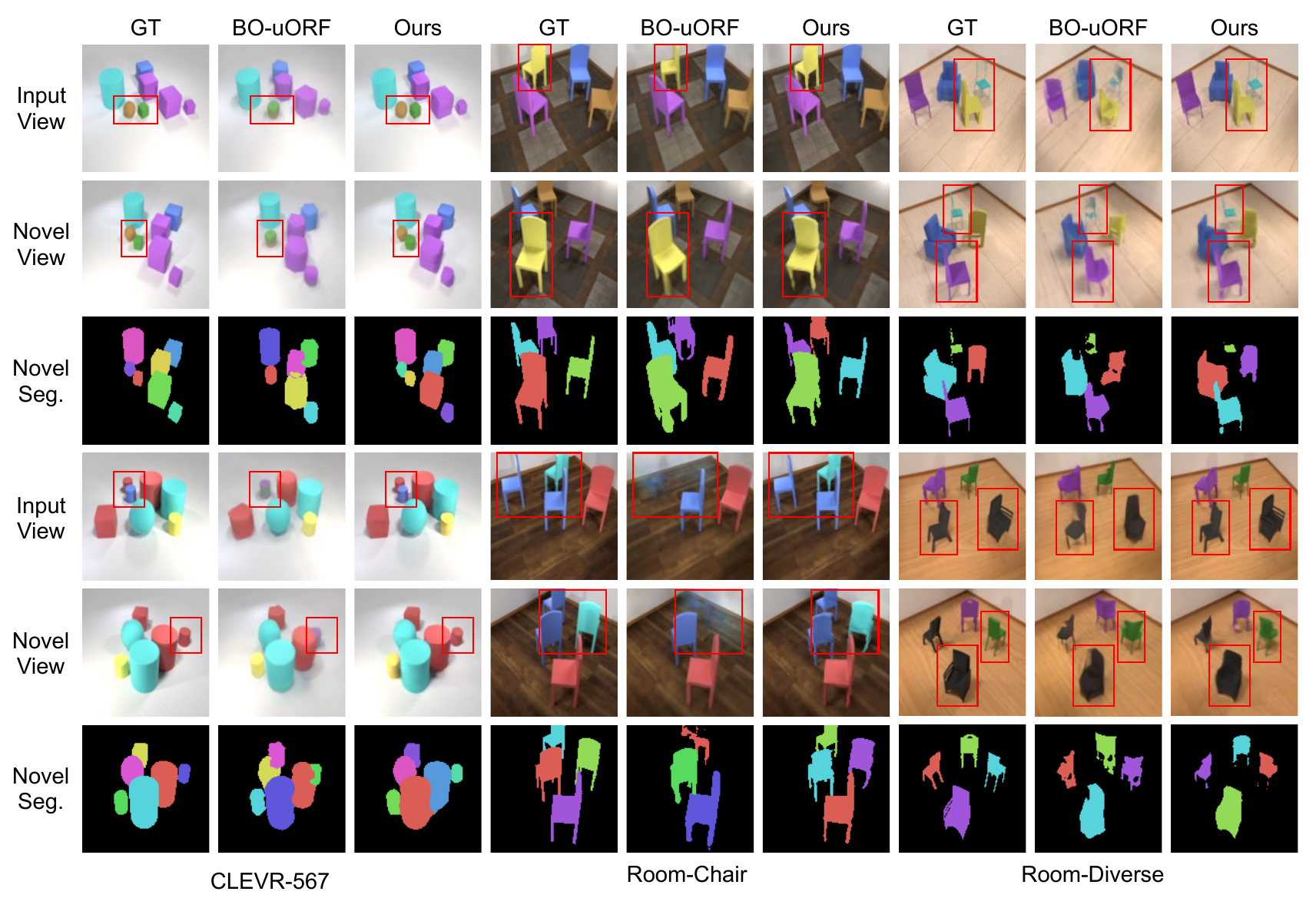}}
    \caption{\textbf{Qualitative comparison on synthetic scenes.} Compared to BO-uORF, \model renders novel-view images and segmentation masks in much higher quality, especially in detailed object attributes like color and shape (best viewed with zoom-in for the {\color{red}{highlighted details}}).}
    \label{fig:qualitative_synthetic}
    \vspace{-.15in}
\end{figure*}

\paragraph{Results and Analysis}
\begin{wrapfigure}[10]{c}{0.5\linewidth}
\raisebox{0pt}[\dimexpr\height-2\baselineskip\relax]{
\begin{minipage}{\linewidth}
\centering
\captionof{table}{\textbf{Quantitative comparison for scene decomposition and novel view synthesis on Room-Texture.}}
\label{tab:room_texture}
\centering
    % \vspace{+2pt}
    \resizebox{\linewidth}{!}{
        \begin{tabular}{ccccccc}
        \toprule
        \multirow{2}[2]{*}{Method} & \multicolumn{3}{c}{Scece segmentation} & \multicolumn{3}{c}{Novel view synthesis}\\
        \cmidrule(lr){2-4}\cmidrule(lr){5-7}
        & NV-ARI$\uparrow$ & ARI$\uparrow$ & FG-ARI$\uparrow$ & LPIPS$\downarrow$ & SSIM$\uparrow$ & PSNR$\uparrow$ \\
        \midrule
        uORF~\cite{yu2021unsupervised} &57.8  &67.0 &9.3 &0.254 &0.711 &24.23 \\
        BO-uORF~\cite{jia2022improving} &60.4  &69.7 &35.4 &0.215 &0.739 &25.26 \\
        COLF~\cite{smith2023colf} &1.1 &23.5 &53.2 &0.504 &0.670 &22.98 \\
        uOCF-N~\cite{luo2024unsupervised} &72.2 &79.1 &58.4 &0.138 &0.796 &28.81 \\
        uOCF-P~\cite{luo2024unsupervised} &70.4 &78.5 &56.3 &0.136 &0.798 &28.85 \\
        \midrule
        \model & \textbf{79.3} & \textbf{86.0} & \textbf{70.7} & \textbf{0.131} &\textbf{0.858} & \textbf{30.68} \\
        \bottomrule
        \end{tabular}
    }
\end{minipage}}
\end{wrapfigure}
We evaluate the performance of \model for unsupervised scene decomposition and present our quantitative results in \cref{tab:synthetic_seg} and \cref{tab:room_texture}. \model outperforms existing 3D object-centric learning methods, achieving the best performance across all datasets. We also visualize qualitative results for segmentation in \cref{fig:qualitative_synthetic} and \cref{fig:qualitative_uocf}. As shown in~\cref{tab:synthetic_seg} and~\cref{tab:room_texture}, \model significantly outperforms current \sota methods by a large margin on all datasets. We also observe from~\cref{fig:qualitative_synthetic} and~\cref{fig:qualitative_uocf} that \model better handles occlusion between objects, offering more complete segmentation. Notably, compared with task-specific auxiliary designs in current baselines (\eg, adversarial loss used in uORF), \model models each slot equivalently and relies solely on the reconstruction loss $\mathcal{L}_{\text{recon}}$ for achieving the good performance. We attribute this effectiveness to our scene encoding design and provide more analyses in~\cref{sec:exp:ablation}.

We also evaluate the capability of our \model for novel-view synthesis and present our quantitative results compared with existing methods in \cref{tab:room_texture}, \cref{tab:synthetic_recon},  and visualize qualitative results in \cref{fig:qualitative_synthetic}, \cref{fig:qualitative_uocf}. As shown in \cref{tab:room_texture} and \cref{tab:synthetic_recon}, our model outperforms existing methods on almost all metrics across the four datasets, rendering novel views of much higher quality, especially for complex datasets. As visualized in \cref{fig:qualitative_synthetic} and \cref{fig:qualitative_uocf}, \model captures more detailed texture, shape, and pose of objects compared with baseline models. 

\begin{table*}[t!]
    \caption{\textbf{Quantitative comparison for novel-view synthesis in synthetic scenes.} \model outperforms existing methods on the majority of metrics in three datasets, rendering novel views of much higher quality, especially for complex datasets.}\label{tab:synthetic_recon}
    \vspace{-8pt}
    \centering
    \small
    \resizebox{\linewidth}{!}{
    \begin{tabular}{cccccccccc}
    \toprule
    \multirow{2}[2]{*}{Method} & \multicolumn{3}{c}{CLEVR-567} & \multicolumn{3}{c}{Room-Chair}& \multicolumn{3}{c}{Room-Diverse}\\
    \cmidrule(lr){2-4}\cmidrule(lr){5-7}\cmidrule(lr){8-10}
    & LPIPS$\downarrow$ & SSIM$\uparrow$ & PSNR$\uparrow$ & LPIPS$\downarrow$ & SSIM$\uparrow$ & PSNR$\uparrow$ & LPIPS$\downarrow$ & SSIM$\uparrow$ & PSNR$\uparrow$ \\
    \midrule
    NeRF-AE~\cite{yu2021unsupervised} & 0.1288 & 0.8658 & 27.16 & 0.1166 & 0.8265 & 28.13 & 0.2458 & 0.6688 & 24.80 \\
    uORF~\cite{yu2021unsupervised} & 0.0859 & 0.8971 & 29.28 & 0.0821 & 0.8722 & 29.60 & 0.1729 & 0.7094 & 25.96 \\
    BO-uORF~\cite{jia2022improving} & 0.0618 & 0.9260 & 30.85 & 0.0733 & 0.8938 & 30.61 & 0.1515 & 0.7363 & 26.96 \\
    COLF~\cite{smith2023colf} & 0.0608 & 0.9346 & 31.81 & 0.0485 & 0.8934 & 30.93 & 0.1274 & 0.7308 & 26.02 \\
    sVORF~\cite{qi2023slot} &0.0211 &\textbf{0.9701} &\textbf{37.20} &0.0824 &0.8992 &33.04 &0.1637 &0.7825 &29.41\\
    \midrule
    \model & \textbf{0.0184} &0.9680 & {36.09} & \textbf{0.0410} & \textbf{0.9358} & \textbf{34.63} & \textbf{0.1159} & \textbf{0.8479} & \textbf{29.97} \\
    \bottomrule
    \end{tabular}
    }
    \vspace{-.15in}
\end{table*}
Additionally, compared to uORF~\cite{yu2021unsupervised} that needs to train for 6 days on Room-Diverse with a single Nvidia RTX 3090 GPU, \model is more efficient, requiring only 30 hours (5x speed up) training time. This is afforded by: (i) the feature lifting design provides detailed information for rendering and leads to a faster model convergence rate; (ii) the slot-based density prediction and rendering in \model requires only 1 radiance field while models like uORF, uOCF, and sVORF compute K fields for each slot; (iii) with no auxiliary losses on the fully rendered image, \model only needs 1024 (or even 256) sampled rays for training with the reconstruction loss, thus largely reducing the computation overhead. Please refer to \cref{app:tab:efficiency} in the \supp for more comparisons.

\subsection{Object-centric Learning in Real-world Scenes}\label{sec:exp:real_render}
\paragraph{Setup} To show the effectiveness of \model on real-world complex scenes, we evaluate \model on Kitchen-Shiny and Kitchen-Matte following uOCF~\cite{luo2024unsupervised}. We use the same train/test split for these two datasets with \textbf{single-view input} following settings in uOCF. Unlike uOCF which requires training with 2 stages to learn object priors, we train \model with reconstruction loss in 1 stage. 

We also consider ScanNet~\cite{dai2017scannet} and DTU~\cite{jensen2014large}, which are well-established datasets for evaluating generalizable novel-view synthesis~\cite{wei2021nerfingmvs, zhang2022nerfusion, gao2023surfelnerf}, as more challenging real-world benchmarks to test models' capability on processing complex real-world scenes. For ScanNet, we follow the standard training and evaluation scheme in existing works~\cite {wei2021nerfingmvs,zhang2022nerfusion}, sample 100 scenes for training, and evaluate our method on the 8 unseen testing scenes introduced. On DTU, we follow the setup of PixelNeRF~\cite{yu2021pixelnerf} and NeRFusion~\cite{zhang2022nerfusion}, train all models on the 88 training scenes, and test on the 15 test scenes. For both ScanNet and DTU, we follow the standard setting in generalizable novel-view synthesis and provide \textbf{4 source nearby views} selected according to previous work~\cite{wang2021ibrnet, zhang2022nerfusion, gao2023surfelnerf, xu2022point, chen2021mvsnerf,varma2022attention} as inputs.

\begin{table*}[t!]
\begin{minipage}{0.55\linewidth}
    \centering
    \caption{\textbf{Quantitative comparison for novel view synthesis on Kitchen-Shiny and Kitchen-Matte.} \model presents significant improvements ($\sim$4 PSNR) and the best results on all perceptual scores.} 
    \label{tab:uocf_data}
    \vspace{-8pt}
    \resizebox{\linewidth}{!}{
        \begin{tabular}{ccccccc}
        \toprule
        \multirow{2}[2]{*}{Method}
        & \multicolumn{3}{c}{Kitchen-Shiny} & \multicolumn{3}{c}{Kitchen-Matte}\\
        \cmidrule(lr){2-4}\cmidrule(lr){5-7}
        & LPIPS$\downarrow$ & SSIM$\uparrow$ & PSNR$\uparrow$ & LPIPS$\downarrow$ & SSIM$\uparrow$ & PSNR$\uparrow$ \\
        \midrule
        uORF~\cite{yu2021unsupervised} &0.336 &0.602 &19.23 &0.092 &0.808 &26.07 \\
        BO-uORF~\cite{jia2022improving} &0.318 &0.639 &19.78 &0.067 &0.832 &27.36 \\
        COLF~\cite{smith2023colf} &0.397 &0.561 &18.30 &0.236 &0.643 &20.68 \\
        uOCF-N~\cite{luo2024unsupervised} &0.055 &0.842 &27.87 &0.055 &0.841 &28.25 \\
        uOCF-P~\cite{luo2024unsupervised} &0.049 &0.862 &28.58 &0.043 &0.867 &29.40 \\
        \midrule
        \model & \textbf{0.035} &\textbf{0.928} &\textbf{32.02} &\textbf{0.030} &\textbf{0.939}  &\textbf{32.92} \\
        \bottomrule
        \end{tabular}
    }
% \vspace{-8pt}
\end{minipage}
\hfill
\begin{minipage}{0.44\linewidth}
\centering
\caption{\textbf{Quantitative comparison on ScanNet.} $\dagger$ We use the official implementations provided to re-train and evaluate the models on ScanNet.}
\label{tab:real_render}
\vspace{-8pt}
    \resizebox{\linewidth}{!}{
        \begin{tabular}{ccccc}
        \toprule
        Method
        % & \multicolumn{3}{c}{DTU MVS}\\
        % & PSNR$\uparrow$ & SSIM$\uparrow$ & LPIPS$\downarrow$
        & PSNR$\uparrow$ & SSIM$\uparrow$ & LPIPS$\downarrow$ &NV-FG-ARI $\uparrow$\\ 
        \midrule
        IBRNet~\cite{wang2021ibrnet} & 21.19 & 0.786 & 0.358 & - \\
        NeRFusion~\cite{zhang2022nerfusion} & 22.99 & 0.838 & 0.335 & - \\
        PointNeRF~\cite{xu2022point} & 20.47 & 0.642 & 0.544 & - \\
        SurfelNeRF~\cite{gao2023surfelnerf} & 23.82 & 0.845 & 0.327 & - \\
        GNT$^\dagger$~\cite{varma2022attention} & 27.76 & 0.8791 & 0.2197 & - \\
        \midrule
        BO-uORF$^\dagger$~\cite{yu2021unsupervised} & 12.72 & 0.3393 & 0.6975 & 0.0 \\
        OSRT$^\dagger$~\cite{sajjadi2022object} &13.34 &0.2746 &0.6337 & 29.7 \\
        \midrule
        \model & \textbf{28.36} & \textbf{0.9200} & \textbf{0.1891} & \textbf{31.1}\\
        \bottomrule
        \end{tabular}
    }
\end{minipage}
\vspace{-.15in}
\end{table*}

\begin{figure*}[t!]
    \centering
    % \fbox{\rule[0cm]{0cm}{12cm}\rule[0cm]{\linewidth}{0cm}}
    \resizebox{\linewidth}{!}{\includegraphics[width=\linewidth]{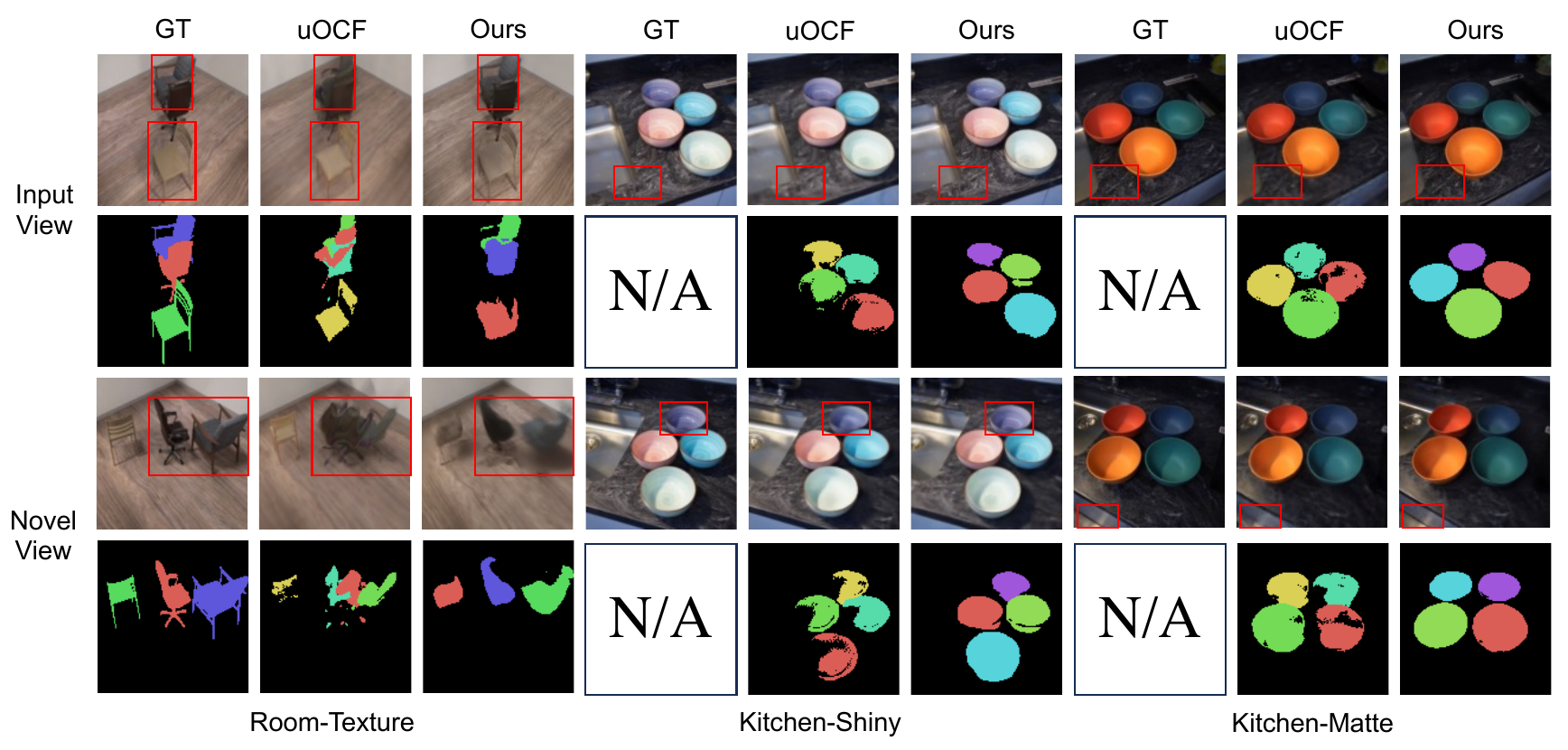}}
    \caption{\textbf{Qualitative comparison on Room-Texture, Kitchen-Shiny, and Kitchen-Matte.} Compared to the SOTA method uOCF, \model renders novel-view images and segmentation masks in higher quality, offering more complete segmentation and more detailed textures (best viewed with zoom-in for the {\color{red}{highlighted details}}). }
    \label{fig:qualitative_uocf}
    \vspace{-.15in}
\end{figure*}

\paragraph{Baselines} For evaluating object-centric learning, we compare our \model with existing state-of-the-art 3D object-centric models, including uORF, BO-uORF, COLF, and uOCF on Kitchen-Shiny and Kitchen-Matte.
On ScanNet, we mainly compare the \model with the improved uORF model for object-centric learning as uOCF requires a two-stage training scheme with auxiliary losses thus not directly comparable. We additionally add OSRT~\cite{sajjadi2022object} as a powerful baseline as it has demonstrated its effectiveness in decomposing complex scenes.

For generalizable novel-view synthesis, compare \model and \sota generalizable NeRFs like NeRFusion~\cite{zhang2022nerfusion} on ScanNet and DTU MVS. 
Additionally, we re-train the recent \sota method GNT~\cite{varma2022attention} for generalizable novel-view synthesis on ScanNet as a strong baseline to validate the effectiveness of our method (see more implementation details in \cref{app:baseline}).

\begin{figure*}[t!]
    \centering
    % \fbox{\rule[0cm]{0cm}{12cm}\rule[0cm]{\linewidth}{0cm}}
    \resizebox{\linewidth}{!}{\includegraphics[width=\linewidth]{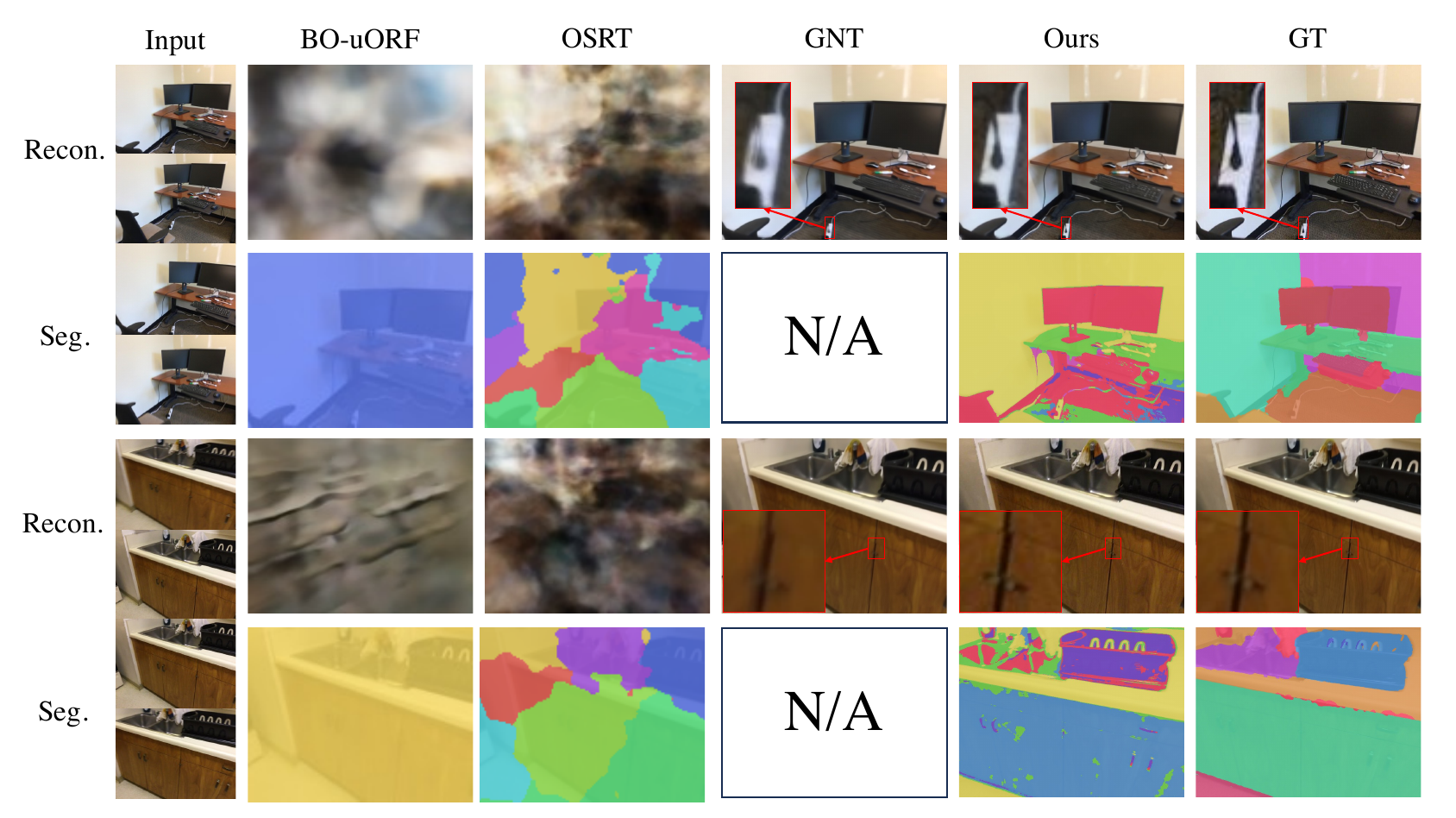}}
    \caption{\textbf{Qualitative results on ScanNet}. Our \model achieves the best performance for novel-view rendering, even surpassing the recent \sota model GNT, while BO-uORF and OSRT struggle to render novel-view images on ScanNet.}
    \label{fig:qualitative_real}
    \vspace{-.15in}
\end{figure*}

\paragraph{Results and Analysis}
We present quantitative evaluations in \cref{tab:uocf_data} and \cref{tab:real_render}, and visualize qualitative results in \cref{fig:qualitative_uocf} and \cref{fig:qualitative_real}. Similar to results in synthetic datasets, we observe a consistent improvement in object-centric learning on real-world datasets. In Kitchen-Shiny and Kitchen-Matte, as there is no ground truth segmentation annotation available, we qualitatively compare \model with existing methods in~\cref{fig:qualitative_uocf}. We demonstrate that \model renders segmentation masks with higher quality, offering more complete object segmentations. The quantitative evaluation results on ScanNet in~\cref{tab:real_render} also demonstrate that \model outperforms existing 3D object-centric methods with more accurate segmentation masks predicted as shown in~\cref{fig:qualitative_real}. Notably,~\cref{fig:qualitative_real} also shows that despite the relatively marginal performance gap (compared with improvements in synthetic datasets) between OSRT and \model in NV-ARI-FG, OSRT generates uniformly distributed masks without properly separating the objects. This originates from an unfair privilege of OSRT when calculating ARI as this metric mainly considers coverage as an important factor. We provide further analyses and discussions on improving \model for complex real-world scenes in \cref{app:sec:potential}.

As shown in~\crefrange{tab:uocf_data}{tab:real_render_dtu}, we observe a consistent advantage of \model on most datasets for novel-view synthesis. This includes outperforming \sota methods dedicatedly designed for generalizable novel-view synthesis like SurfelNeRF and GNT. Meanwhile,~\cref{tab:real_render} and~\cref{fig:qualitative_real} show that methods like BO-uORF and OSRT struggle to render novel-view images in complex settings, achieving only a PSNR of less than 14 with no meaningful rendered results. Notably, OSRT achieves a PSNR of 27 on training scenes but fails to generalize to unseen scenes (see more discussions in \cref{app:discuss}). These results further validate the effectiveness of \model compared with previous 3D object-centric learning methods.

\begin{table*}[t!]
\begin{minipage}{0.5\linewidth}
    \caption{\textbf{Quantitative comparison on DTU.}}
    \label{tab:real_render_dtu}
    \vspace{-8pt}
    \centering
    \resizebox{0.9\linewidth}{!}{
        \begin{tabular}{cccc}
        \toprule
        Method
        & PSNR$\uparrow$ & SSIM$\uparrow$ & LPIPS$\downarrow$\\ 
        \midrule
        PixelNeRF~\cite{yu2021pixelnerf} &19.31 &0.789 &0.382 \\
        IBRNet~\cite{wang2021ibrnet} & 26.04 & 0.917 & 0.190 \\
        MVSNeRF~\cite{chen2021mvsnerf} &26.63 &\textbf{0.931} &0.168 \\
        NeRFusion~\cite{zhang2022nerfusion} & 26.19 & 0.922 & 0.177 \\
        \midrule
        \model & \textbf{26.75} & 0.896 & \textbf{0.157} \\
        \bottomrule
        \end{tabular}
    }
\end{minipage}
\begin{minipage}{0.48\linewidth}
    \centering
    \caption{\textbf{Sensitivity of random masking ratio scheduling.}}\label{tab:albation_mask}
    \vspace{-8pt}
    \resizebox{\linewidth}{!}{
        \begin{tabular}{ccccc}
        \toprule
        Decay Steps & PSNR$\uparrow$ & NV-ARI$\uparrow$ & ARI$\uparrow$ &FG-ARI $\uparrow$\\ 
        \midrule
        0 & 29.89 & 74.4 & 85.8 & 43.6 \\
        % 3000 &13.34 &0.2746 &0.6337 & 29.7 \\
        10000 &\textbf{30.01} &74.9 &86.9 & 42.1 \\
        30000 &29.80 &\textbf{77.5} &\textbf{90.3} & 84.8 \\
        60000 &29.53 &77.3 &90.1 & \textbf{85.7} \\
        100000 &28.68 &76.4 &89.4 &83.6 \\
        \bottomrule
        \end{tabular}
    }
\end{minipage}
\vspace{-.15in}
\end{table*}

\subsection{Ablative Study}\label{sec:exp:ablation}
To investigate the effectiveness of our designs in \model, including scene encoding, random masking, slot-based density, and the number of slots and source views, we conduct ablative studies on both synthetic (Room-Diverse) and real-world (ScanNet) scenes. We also investigate the effect of the number of sampled rays and leave the results in \cref{app:tab:ablation_rays} in the supplementary.

\begin{table*}[t!]
    \centering
    \caption{\textbf{Ablations analysis of module designs in \model.} } \label{tab:ablation}
    \vspace{-8pt}
    \resizebox{\linewidth}{!}{
    \begin{tabular}{ccccccccccc}
    \toprule
    \multirow{2}[2]{*}{Method} & \multicolumn{6}{c}{Room-Diverse} & \multicolumn{4}{c}{ScanNet}\\
    \cmidrule(lr){2-7}\cmidrule(lr){8-11}
    & LPIPS$\downarrow$ & SSIM$\uparrow$ & PSNR$\uparrow$ & NV-ARI$\uparrow$ & ARI$\uparrow$ & FG-ARI$\uparrow$ &LPIPS$\downarrow$ & SSIM$\uparrow$ & PSNR$\uparrow$& NV-FG-ARI$\uparrow$ \\
    \midrule
    w/o Feature Lift.  &0.2537  &0.7716  &28.20  &71.4  &75.8  &65.3  &0.5622  &0.5129  &11.60 &0.0 \\
    w/o Random Mask  &\textbf{0.1169}  &\textbf{0.8470}  &\textbf{29.89}  &74.4  &85.8  &43.6  &\textbf{0.1861} &\textbf{0.9208} &27.86 & 17.63\\
    w/o Slot Density  &0.1180  &0.8456  &29.82  &76.3  &87.6  &77.3 &0.1937 &0.9134  &27.42  &6.6 \\
    \midrule
    FullModel & 0.1180 & 0.8454 & 29.80 & \textbf{77.5} & \textbf{90.3} & \textbf{84.8} &0.1891 &0.9200 &\textbf{28.36} & \textbf{31.1}\\
    \bottomrule
    \end{tabular}
    }
    \vspace{-.15in}
\end{table*}

\begin{figure}[t!]
\begin{minipage}{0.46\linewidth}
    \centering
    \resizebox{\linewidth}{!}{\includegraphics[width=\linewidth]{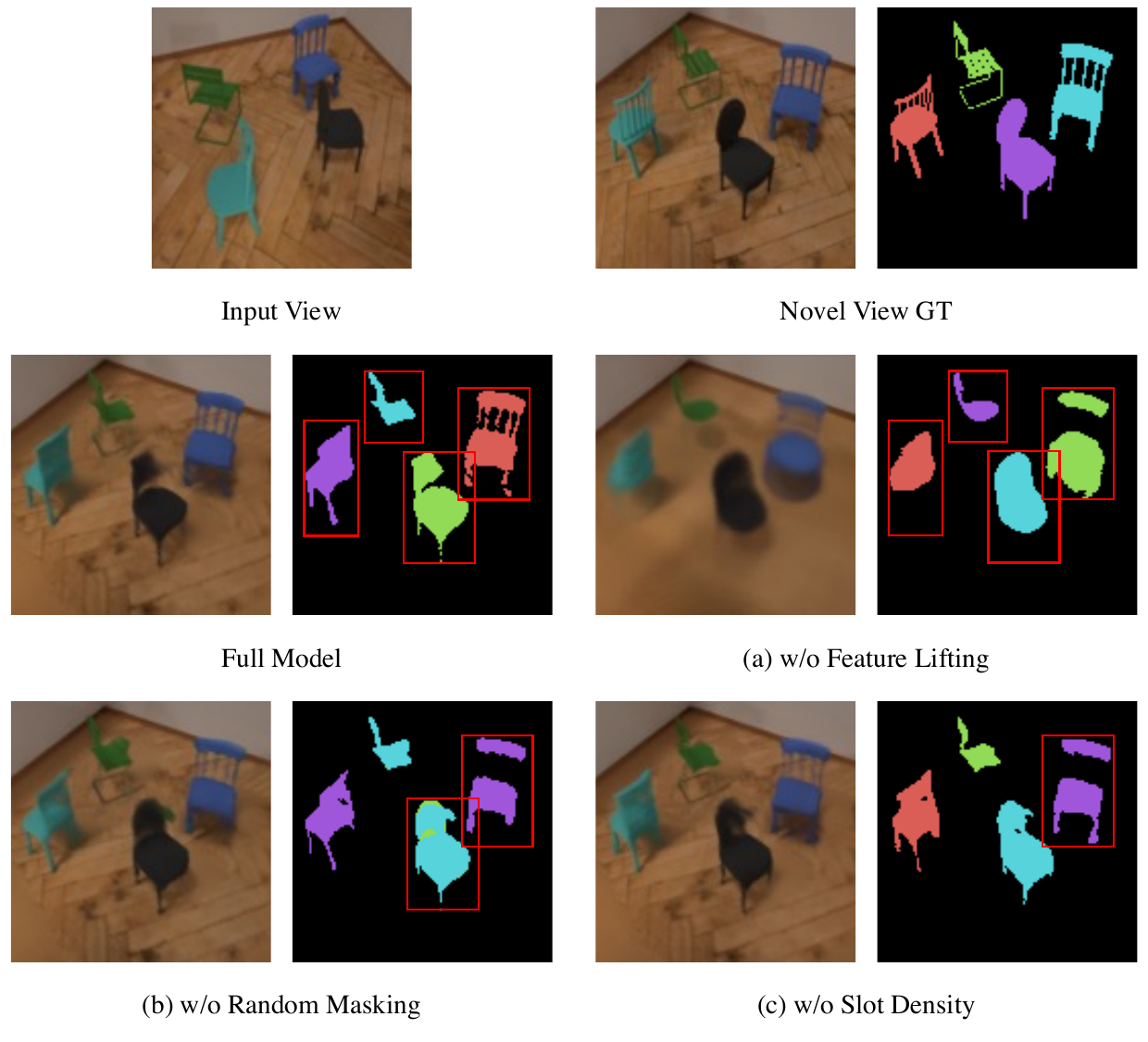}}
    \caption{\textbf{Visualization of model ablation analysis.} (a) Without feature lifting, \model renders blurred images and imprecise segmentation masks. (b) Without random masking, \model cannot segment objects correctly. (c) Using slot-based density helps \model learn more accurate segmentation.}
    \label{fig:ablation}
\end{minipage}
\hfill
\begin{minipage}{0.51\linewidth}
    \centering
    \includegraphics[width=\linewidth]{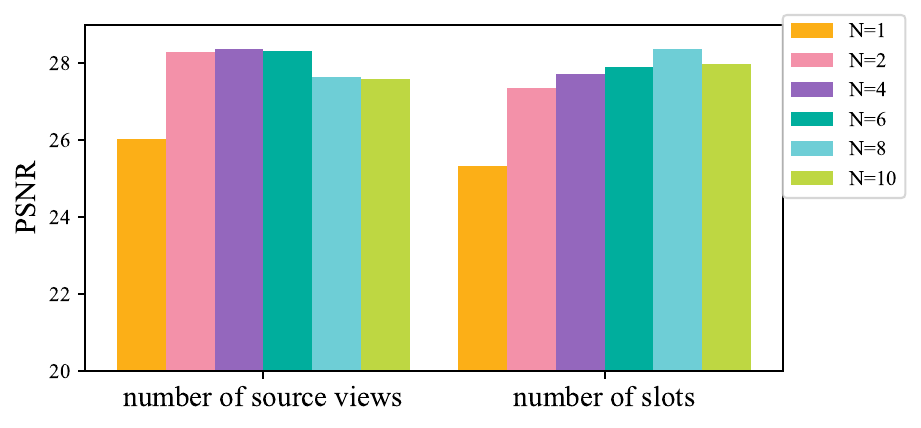}
    \includegraphics[width=\linewidth]{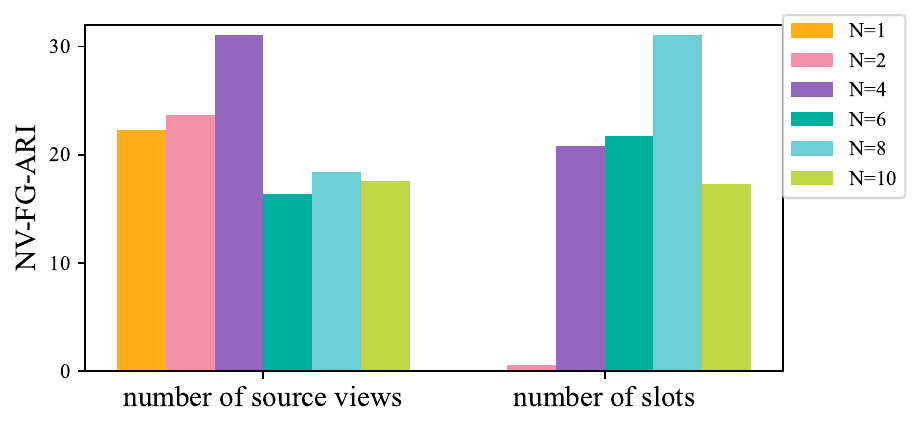}
    \caption{\textbf{Ablative studies over the number of source views and slots.} We set the number of slots to 8 for different numbers of source views and set the number of source views to 4 for different numbers of slots.}
    \label{fig:ablation_num}
\end{minipage}
\vspace{-.15in}
\end{figure}

\paragraph{Scene Encoding} We consider removing the feature lifting operation and initializing point features solely with positional embeddings, \ie, $\mF_p=\mE_p$. As shown in~\cref{tab:ablation} and \cref{fig:ablation}, the performance of both novel-view synthesis and scene decomposition on Room-Diverse drops significantly without lifted multi-view features, especially for LPIPS and FG-ARI. In fact, it is hard to establish the mapping between slots and 3D points via only positional information. This problem is more severe in complex real-world scenes (\eg, Scannet), where \model struggles in rendering novel views without feature lifting, achieving only a PSNR of 11.6. This issue is also shared by uORF and OSRT as presented in~\cref{sec:exp:real_render} and demonstrates the significance of the feature lifting design.

\paragraph{Random Masking} As shown in \cref{tab:ablation} and \cref{fig:ablation}, abandoning the random masking scheme described in \cref{sec:method:training} slightly improves the rendering performance (LPIPS, SSIM, PSNR) but significantly decreases the scene decomposition capability of \model, especially for FG-ARI. We also find that the model sometimes converges to the degenerate scenario as discussed in \cref{sec:method:training} without the random masking scheme, leading to a collapse in scene decomposition (\ie, uniform segmentation predictions) with ARI scores lower than 40. This affirms our supposition that, without random masking, the model is likely to degenerate and rely solely on lifted features for rendering, thereby ignoring the information in slots. We also explore how the masking ratio decay scheduling influences performance. As shown in \cref{tab:albation_mask}, increasing decay steps slightly harms rendering performance and significantly improves segmentation performance after a certain amount of steps ($\sim$10K steps). After the number of decay steps exceeds 30K, continuing to increase the number of steps will only bring marginal improvement. 

\paragraph{Slot-based Density} As shown in \cref{tab:ablation}, compared with using an additional MLP layer for predicting the density value, using slot-based density slightly improves the quality of novel-view synthesis on ScanNet and significantly improves the performance of scene decomposition on both datasets, especially for ScanNet. We attribute this effectiveness to the fact that the slot-based density is more involved in point-slot interactions. This leads to more information propagation to slots, thus improving the learned object-centric representations for accurately segmenting foreground objects.

\paragraph{Sensitivity to Number of Slots and Source Views} As discussed in~\cref{sec:method:slot_lifting}, \model can accept a various number of source views as input. We investigate how the number of slots and source views influences the performance of \model on ScanNet. As shown in~\cref{fig:ablation_num}, \model is sensitive to the number of slots, which is consistent with previous research on \sa. In addition, the number of source views also has a significant impact on model performance, as it influences both the extracted slots and the lifted 3D point features which are essential components for slot-guided feature lifting in \model.

\section{Conclusion}
We present \model, an object-centric radiance field model for unsupervised 3D object-centric representation learning. Our \model employs slot-guided feature lifting to improve the interaction between lifted input view features and learned slots during decoding. \model achieves \sota performance with large improvements on four challenging synthetic and four complex real-world datasets for scene decomposition and novel-view synthesis and uses much less training time, demonstrating its effectiveness and efficiency. Furthermore, \model demonstrates superior performance for novel-view synthesis on real-world datasets, underscoring its potential to narrow the gap to real-world scenes.

% \clearpage  % TODO REVIEW/FINAL: This \clearpage needs to be removed from both review and camera-ready versions.

\section*{Acknowledgement}
We gratefully thank all colleagues from BIGAI for fruitful discussions. We would also like to thank the anonymous reviewers for their constructive feedback. This work reported herein was supported by Beijing Natural Science Foundation (QY23126).
\bibliographystyle{splncs04}
% \bibliography{main}

\clearpage
\appendix
\setcounter{page}{1}
\renewcommand{\thefigure}{A.\arabic{figure}}
\renewcommand{\thetable}{A.\arabic{table}}
\renewcommand{\theequation}{A.\arabic{equation}}
\setcounter{section}{0}
\setcounter{figure}{0}
\setcounter{table}{0}
\setcounter{equation}{0}
% \maketitlesupplementary

{
    \Large
    \begin{center}
    \textbf{\textsc{SlotLifter}: Slot-guided Feature Lifting for } \\
    \textbf{Learning Object-centric Radiance Fields} \\ 
    \vspace{0.5em}
    Supplementary Material
    \vspace{0.5em}
    \end{center}
}

\section{Implementation Details}\label{app:imp}
\subsection{\model}
\subsubsection{Architecture Design}
\paragraph{Scene Encoding} We employ a U-net-like encoder $E_\Phi$ with ResNet34~\cite{he2016deep} to extract 2D image features, similar to IBRNet~\cite{wang2021ibrnet}. This architecture truncates after \texttt{layer3} as the encoder and adds two up-sampling layers with convolutions and skip-connections as the decoder. Instead of extracting two sets of feature maps for coarse and fine networks as IBRNet, we extract a shared feature map. In addition, we concatenate multi-view images with their corresponding ray directions and camera positions to provide more spatial information, enabling slots to learn 3D information from 2D multi-view features via \sa. Given extracted feature maps, we obtain slots via \sa and 3D point features via feature lifting described in~\cref{sec:method:slot_lifting}. We add the point positional embedding $\mE_p$ in~\cref{eq:fusion}, which considers point location $\vp$ and ray direction $\vd$ simultaneously by:
\begin{equation*}
    \mE_p = \mathrm{MLP}(\mathrm{Concat}([\mathrm{PosEmb}(\vp),\mathrm{PosEmb}(\vd)])),
\end{equation*}
where PosEmb is a Fourier transformation with a frequency of 10 while MLP is used to fuse point location and ray direction information and project positional embedding to the same dimension as point features.
\paragraph{Point-slot Joint Decoding} Our point-slot joint decoding contains an allocation transformer and an attention-based point-slot mapping module. The allocation transformer consists of four transformer layers, and each layer includes a cross-attention layer, a 1D convolution layer, and a self-attention layer. We use 1D convolution and self-attention to model the relationship among points along a ray. The design is based on the insight that spatially adjacent points are more likely to be associated with the same slot. Additionally, We use the weighted sum of attention weights to estimate the density value in~\cref{equ:density}. As this design may restrict the scale of density by the attention weights between slots and point features, the density obtained from~\cref{equ:density} is multiplied by a learnable parameter $s_\sigma$ to rescale it.

\begin{table}[t!]
% \begin{minipage}{0.6\linewidth}
\caption{Training configuration for our \model. The values in parentheses are adopted for the ScanNet and DTU datasets.} \label{tab:config}
\centering
\resizebox{0.6\linewidth}{!}{
    \begin{tabular}{ccc}
        \toprule
        \multirow{8}{*}{Training} & Scene Batch Size &4 (2) \\
        & Ray Batch Size & 1024 \\
        & LR  &5e-5 \\
        & LR Warm-up Steps  &10000 \\
        & LR Decay Steps  &50000 \\
        & Max Steps  &250K \\
        & Num. Source Views &1 (4) \\
        & Grad. Clip &0.5 \\
        \midrule
        \multirow{4}{*}{Scene Encoding} &Feature Dimension &64 (32) \\
        & Slot Dimension &256 \\
        &Iterations &3 \\
        & $\sigma$ Annealing Steps &30000 \\
        \midrule
        \multirow{3}{*}{Point-slot Decoding} &Num. Layers &4 \\
        &Attention Heads &4 \\
        &Feature Dimension &64 \\
        \bottomrule
    \end{tabular}
}
% \end{minipage}
\end{table}

\begin{table}[t!]
\caption{Image resolution and the number of slots on different datasets.} \label{tab:num_slots}
\centering
\resizebox{\linewidth}{!}{
    \begin{tabular}{ccccc}
        \toprule
        Dataset &CLEVR567 &Room-Chair &Room-Diverse &Room-Texture\\
        \midrule
        Resolution &128$\times$128 &128$\times$128 &128$\times$128 &128$\times$128 \\
        Number of slots &8 &5 &5 &5  \\
        \midrule
        Dataset &Kitchen-Shiny &Kitchen-Matte &DTU MVS &ScanNet\\
        \midrule
        Resolution &128$\times$128 &128$\times$128 &400$\times$300 &640$\times$480 \\
        Number of slots &5 &5 &8 &8 \\
        \bottomrule
    \end{tabular}
}
\end{table}

\subsubsection{Hyperparameters and Training Details}
We train our \model by sampling 1024 rays for each scene with a learning rate of $5\times10^{-5}$, a linear learning rate warm-up of 10000 steps, and an exponentially decaying schedule. All the images are resized to 128$\times$128 for synthetic data and 640$\times$480 for real-world data. Image resolution and the number of slots K used on different datasets are shown in~\cref{tab:num_slots}. To encourage \model to segment the background properly, we use the locality constraint proposed by uORF~\cite{yu2021unsupervised}. Specifically, we set a background bound and enforce every point outside the bound being mapped to the empty slot or the first slot. The locality constraint is imposed for the first 50K iterations, preventing \model from segmenting the background as 2 separate objects. Note that our \model does not require a background-aware \sa like uORF since our slots are initialized by learnable queries, enabling \sa to learn to individually segment the background and foreground objects. On ScanNet and DTU MVS, we adopt the coarse-to-fine sampling scheme on ScanNet following previous methods, sampling 64 points along each ray for the coarse sampling and another 64 points for the fine sampling. We found that the coarse-to-fine sampling scheme aids \model in rendering novel views with higher quality. The training configuration is summarized  in~\cref{tab:config}. Additionally, we found the background occlusion regularization loss from~\cite{luo2024unsupervised, yang2023freenerf} is helpful on the Kitchen-Matte and Kitchen-Shiny datasets for preventing the background slot segmenting foreground objects but it has little effect on the rendering quality. We only use this loss on the Kitchen-Matte and Kitchen-Shiny datasets because we didn't find it helpful on other datasets. 

\subsection{Baselines}
\label{app:baseline}
\subsubsection{uORF and BO-uORF}
The experimental results of uORF~\cite{yu2021unsupervised} on CLEVR-567, Room-Chair, and Room-Diverse are taken from their paper. We trained the BO-uORF model on CLEVR-567, Room-Chair, Room-Diverse, and ScanNet using the official implementation of \href{https://github.com/KovenYu/uORF}{uORF} and \href{https://github.com/YuLiu-LY/BO-QSA}{BO-QSA}. As (BO-)uORF only accepts single source view input, we selected the closest view to the target view as the source view for it. Unfortunately, due to design limitations, such as model architecture, adversarial loss, perceptual loss, etc., we could not train the BO-uORF model at the resolution of 640$\times$480. Therefore, we had to use a resolution of 128$\times$128 following their original settings. We use 8 slots for uORF as same as our method.

\begin{table*}[t!]
    \caption{\textbf{Efficiency and performance comparisons on Room-Diverse.} We evaluate all the methods on an NVIDIA RTX 3090 GPU.} 
    \centering
    \label{app:tab:efficiency}
\resizebox{\linewidth}{!}{
            \begin{tabular}{ccccccc}
            \toprule
            Model & PSNR$\uparrow$ & LPIPS$\downarrow$ & NV-ARI$\uparrow$ & ARI$\uparrow$ & GPU Memory$\downarrow$ & Training Time$\downarrow$ \\
            \midrule
            uORF &25.96 &0.1729 &56.9 &65.6 &24 GB &~6 days \\
            BO-uORF &26.96 &0.1515 &62.5 &72.6 &24 GB &~6 days \\
            ours(N=256) &29.83 &0.1345 &76.1 &88.7 &3.5 GB &10 hours \\
            ours(N=512) &\textbf{29.84} &0.1277 &76.2 &88.0 &6 GB &19 hours \\
            ours(N=1024) &29.80 &\textbf{0.1180} &\textbf{77.5} &\textbf{90.3} &12 GB & 30 hours \\
            \bottomrule
            \end{tabular}
        }
\end{table*}

\begin{table*}[t!]
    \caption{\textbf{Ablations on the number of rays with different image sizes.} Increasing the number of rays sightly improves rendering and segmentation quality, while reducing image size slightly decreases both rendering and segmentation quality.} 
    \centering
    \label{app:tab:ablation_rays}
\resizebox{\linewidth}{!}{
            \begin{tabular}{ccccccccccc}
            \toprule
            \multirow{2}[2]{*}{Number of rays} &\multicolumn{2}{c}{ScanNet (640$\times$480)} &\multicolumn{4}{c}{Room-Diverse (128$\times$128)} &\multicolumn{4}{c}{Room-Diverse (64$\times$64)}\\
            \cmidrule(lr){2-3}\cmidrule(lr){4-7}\cmidrule(lr){8-11}
            & PSNR$\uparrow$ & NV-FG-ARI$\uparrow$ & PSNR$\uparrow$ & LPIPS$\downarrow$ & NV-ARI$\uparrow$ & ARI$\uparrow$ & PSNR$\uparrow$ & LPIPS$\downarrow$ & NV-ARI$\uparrow$ & ARI$\uparrow$\\ 
            \midrule
            256 &27.27 &19.8 &29.83 &0.1345 &76.1 &88.7 &29.55 &0.0792 &69.2 &82.4 \\
            512 &27.92 &\textbf{32.3} &\textbf{29.84} &0.1277 &76.2 &88.0 &\textbf{29.57} &0.0731 &69.9 &\textbf{83.2} \\
            1024 &\textbf{28.36} &31.1 &29.80 &\textbf{0.1180} &\textbf{77.5} &\textbf{90.3} &\textbf{29.57} &\textbf{0.0687} &\textbf{70.2} &82.4 \\
            \bottomrule
            \end{tabular}
        }
\end{table*}

\subsubsection{OSRT}
We trained OSRT~\cite{sajjadi2022object} on CLEVR-567, Room-Chair, Room-Diverse, and ScanNet using the \href{https://github.com/stelzner/osrt}{implementation} recommended by the authors on the \href{https://osrt-paper.github.io/#code}{project website} of OSRT. We observed that OSRT's performance of scene decomposition is highly sensitive to the batch size used during training, which is also mentioned in the \href{https://github.com/stelzner/osrt}{implementation}. Due to the computational limitations, we trained the OSRT for 250K iterations using a batch size of 64 and sampling 2048 rays for each scene with 2 Nvidia A100 GPUs. To train OSRT on the ScanNet dataset, we resized all images to 128$\times$128.
\subsubsection{GNT}
We trained GNT~\cite {varma2022attention} on ScanNet using their official \href{https://github.com/VITA-Group/GNT}{implementation}. We trained GNT for 250K iterations with their config \texttt{gnt\_full.txt} in their repository, which uses a learning rate of $5\times10^{-4}$, samples 2048 rays for each scene, and selects 10 nearby source views to render the target view.

\section{Additional Discussions}
\subsection{Potential Improvements}\label{app:sec:potential}
Although \model exhibits superior performance in novel-view synthesis and scene decomposition compared to \sota 3D object-centric learning methods, its scene decomposition performance still falls short under real-world settings. This is particularly noteworthy considering the recent success of 2D object-centric models on real-world images (see in~\cref{tab:seg_scannet}). We attribute this undesired effect to the unconstrained point-slot mapping process. As elaborated in~\cref{sec:method:slot_lifting}, the slots are mapped to the 3D points which are later projected to the target view image. With only reconstruction loss, the information in the target image can be backpropagated to both slots and lifted point features. This adds no direct guidance or constraints on slot learning and can easily make the learned slots attend to features that best render the scene instead of decomposing it. 

\begin{wrapfigure}[9]{c}{0.5\linewidth}
\raisebox{0pt}[\dimexpr\height-2\baselineskip\relax]{\begin{minipage}{\linewidth}
\centering
\captionof{table}{\textbf{Quantitative segmentation results on ScanNet.} FG-ARI is evaluated on the input view(s). ``MV'' indicates 3D multi-view inputs. }\label{tab:seg_scannet}
\vspace{+2pt}
\resizebox{\linewidth}{!}{
    \begin{tabular}{ccccc}
    \toprule
    Model& Modality &FG-ARI$\uparrow$ &NV-FG-ARI$\uparrow$ &PSNR$\uparrow$ \\
    \midrule
    \sa~\cite{locatello2020sa} & 2D & 31.1 &- &-\\
    DINOSAUR~\cite{seitzer2023bridging} & 2D & \textbf{47.6} &- &-\\
    \midrule
    OSRT~\cite{sajjadi2022object} & MV & 29.8 &29.7 &13.34\\
    \model & MV & 32.0 &31.1 &\textbf{28.36} \\
    \model w/ $\mathcal{L}_{\mathrm{feat}}$ & MV & 36.1 &\textbf{35.7} &25.38 \\
    \bottomrule
    \end{tabular}
}
\end{minipage}}
\end{wrapfigure}

To account for this issue we considered guiding slots to decompose scenes with semantic priors in pre-trained models. Inspired by recent object-centric learning methods DINOSAUR~\cite{seitzer2023bridging} and VideoSAUR~\cite{zadaianchuk2023objectcentric} that replace image reconstruction with feature reconstruction, we propose to improve the scene decomposition capability of \model by adding a feature reconstruction loss. Specifically, we first extract DINOv2~\cite{oquab2023dinov2} features $\hat{\mH}$ for the target view as ground truth. Similar to the color prediction in~\cref{equ:rendering} we add an MLP to predict a feature grid $\vh$ and render 2D features $\mH$. Next, we add the feature reconstruction loss over the predicted target-view feature, \ie $\mathcal{L}_{\mathrm{feat}}=1-D(\mH, \hat{\mH})$, where $D$ denotes the cosine similarity. As shown in~\cref{tab:seg_scannet}, this feature reconstruction loss improves the segmentation performance on ScanNet, but it harms the rendering quality of novel view images, decreasing the PSNR to 25.38. This result reveals the key conflict between the high-level semantic guidance and the low-level appearance guidance which is commonly shared in object-centric models. Adding 3D geometry or temporal constraints (\eg, shape and temporal consistency) that reveal objectness can potentially solve this problem and we leave it as an important future work. 

On the other hand, the superior performance on ScanNet and DTU implies better scene encoding in \model, supporting potential conjectures that these latent slots work similarly to latent feature grids with point features interpolated over them for better novel-view synthesis. This echoes the success of feature-grid-based methods (\eg, Plenoxels~\cite{fridovich2022plenoxels}) for improving the performance of NeRF.

\subsection{Further Discussions about Previous Methods}\label{app:discuss}
\paragraph{(BO-)uORF} As shown in~\cref{tab:real_render} and~\cref{fig:qualitative_real}, BO-uORF failed to render novel views and decompose scenes in complex real-world scenes, achieving only a PSNR of 12.72 and a NV-FG-ARI of 0.0. Moreover, to demonstrate that the failure of BO-uORF is not due to lower resolution, we trained our \model with a resolution of 128$\times$128 and achieved a PSNR of 29.31.
\begin{table*}[t!]
    \caption{\textbf{Quantitative results of OSRT in synthetic scenes.} We present the best performance of our reimplemented OSRT. The performance of OSRT is hindered by its requirements for large amounts of data and computational demands. } 
    \centering
    \label{tab:osrt}
    \resizebox{\linewidth}{!}{
    \begin{tabular}{cccccccccc}
    \toprule
    \multirow{2}[2]{*}{Model} & \multicolumn{3}{c}{CLEVR-567} & \multicolumn{3}{c}{Room-Chair}& \multicolumn{3}{c}{Room-Diverse}\\
    \cmidrule(lr){2-4}\cmidrule(lr){5-7}\cmidrule(lr){8-10}
    & NV-ARI$\uparrow$  & FG-ARI$\uparrow$ &PSNR$\uparrow$  & NV-ARI$\uparrow$  & FG-ARI$\uparrow$ &PSNR$\uparrow$ & NV-ARI$\uparrow$  & FG-ARI$\uparrow$ &PSNR$\uparrow$ \\
    \midrule
    OSRT$^\dagger$~\cite{sajjadi2022object} &3.1  &10.3  &20.73  &5.4  &24.0  &20.99  &7.4 &39.3  &24.58 \\
    uORF~\cite{yu2021unsupervised} & {83.8$\pm$0.3} & 87.4$\pm$0.8 & 29.28 & 74.3$\pm$1.9 & 88.8$\pm$2.7 & 29.60 & 56.9$\pm$0.2 & 67.8$\pm$1.7 & 25.96\\
    \midrule
    \model & \textbf{87.0$\pm$2.8} & \textbf{91.3$\pm$1.8} & \textbf{36.09} & \textbf{89.7$\pm$0.5} & \textbf{91.9$\pm$0.3} & \textbf{34.63} & \textbf{77.5$\pm$0.7} & \textbf{84.3$\pm$2.9} & \textbf{29.97} \\
    \bottomrule
    \end{tabular}
    }
\end{table*}
\begin{figure*}[t!]
    \centering
    \resizebox{\linewidth}{!}{\includegraphics[width=\linewidth]{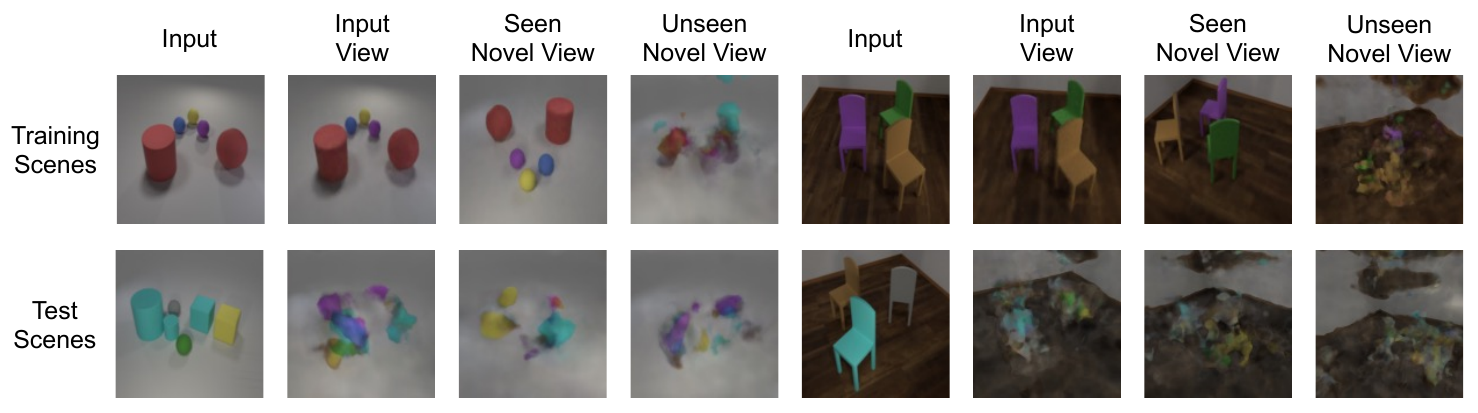}}
    % \resizebox{\linewidth}{!}{
    %     \fbox{\rule[0cm]{0cm}{7cm}\rule[0cm]{\linewidth}{0cm}}
    % }
    \caption{\textbf{Qualitative results of OSRT}. OSRT tends to overfit training scenes and training views, making it difficult to generalize to unseen scenes and unseen views.}
    \label{fig:vis_osrt}
\end{figure*}
\begin{figure*}[t!]
    \centering
    \resizebox{\linewidth}{!}{\includegraphics[width=\linewidth]{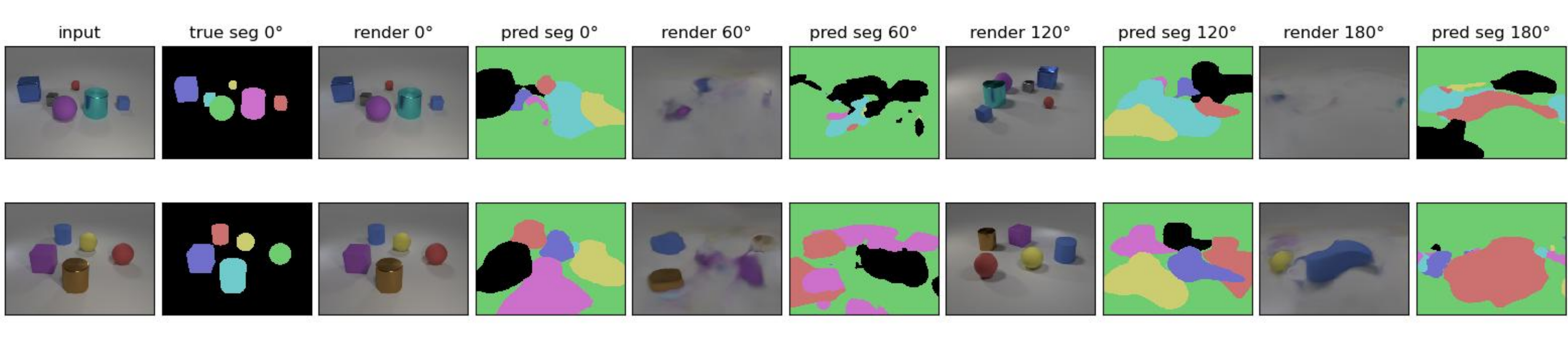}}
    \caption{\textbf{Qualitative results of OSRT from the \href{https://github.com/stelzner/osrt}{implementation} recommended by the authors of OSRT}. OSRT performs well on seen views (\eg, 0$^\circ$, 120$^\circ$), but has difficulty in unseen views (\eg, 60$^\circ$, 180$^\circ$).}
    \label{fig:vis_osrt_clevr}
\end{figure*}
\paragraph{OSRT} We present the quantitative results of OSRT in~\cref{tab:osrt} and visualize qualitative results in~\cref{fig:vis_osrt}. The performance of OSRT is hindered by its requirements for large amounts of data and computational demands, especially on CLEVR-567 which only has 1000 training scenes. We observed that the OSRT tended to overfit training scenes (~\cref{fig:vis_osrt} ), making it difficult to generalize to unseen scenes. We attempted a larger batch size of 256 and trained OSRT for more iterations (750K), but the overfitting issue persisted. We also visualize the results provided by this \href{https://github.com/stelzner/osrt}{implementation} in~\cref{fig:vis_osrt_clevr}, which demonstrates a similar experimental phenomenon that OSRT performs well on seen views (\eg, 0$^\circ$, 120$^\circ$), but has difficulty in unseen views (\eg, 60$^\circ$, 180$^\circ$). Quantitatively, on the CLEVR-567 dataset, OSRT achieved 47+ PSNR on training scenes, but only 20.73 on test scenes. Similarly, on the ScanNet dataset, OSRT achieved 27+ PSNR on training scenes, but only 13.34 on test scenes. These results demonstrate that OSRT may memorize all the training scenes with its powerful transformer encoder-decoder, requiring a lot of data to overcome the overfitting problem. The number of training scenes used in our paper might not be sufficient to train the OSRT(1000 for CLEVR-567 and Room-Chair, 5000 for Room-Diverse, and 100 for ScanNet), leading to the failure case. We attribute this ineffectiveness to its lack of inductive bias for 3D scenes, which is a main distinction between OSRT and our \model.

% \subsection{Per-scene Optimization on ScanNet}
\section{Limitations and Future Work}\label{app:limit}
\begin{figure*}[t!]
    \centering
    \resizebox{\linewidth}{!}{\includegraphics[width=\linewidth]{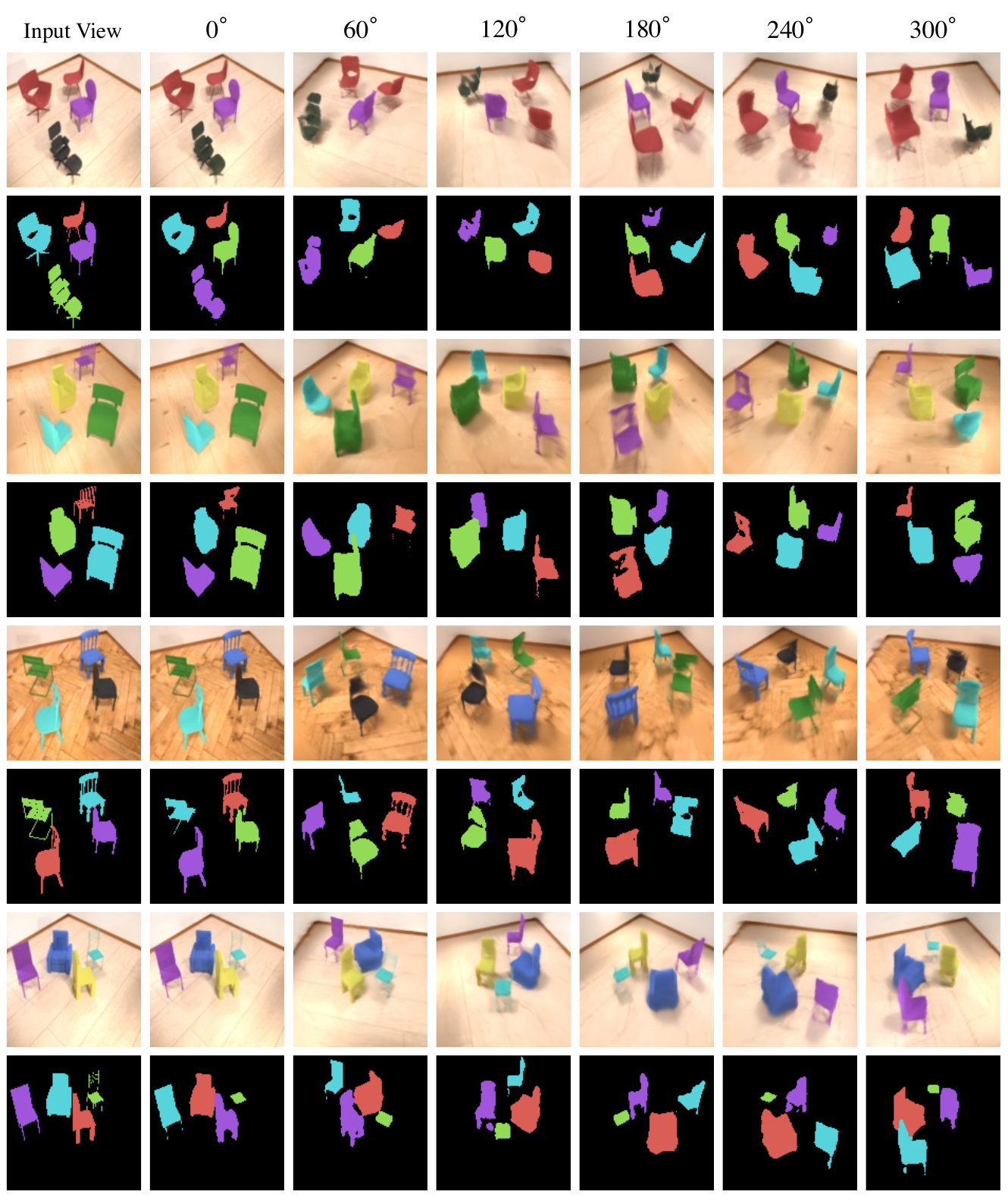}}
    \caption{Novel view synthesis and unsupervised segmentation on Room-Diverse.}
    \label{fig:supp_diverse}
\end{figure*}

\begin{figure*}[t!]
    \centering
    \resizebox{\linewidth}{!}{\includegraphics[width=\linewidth]{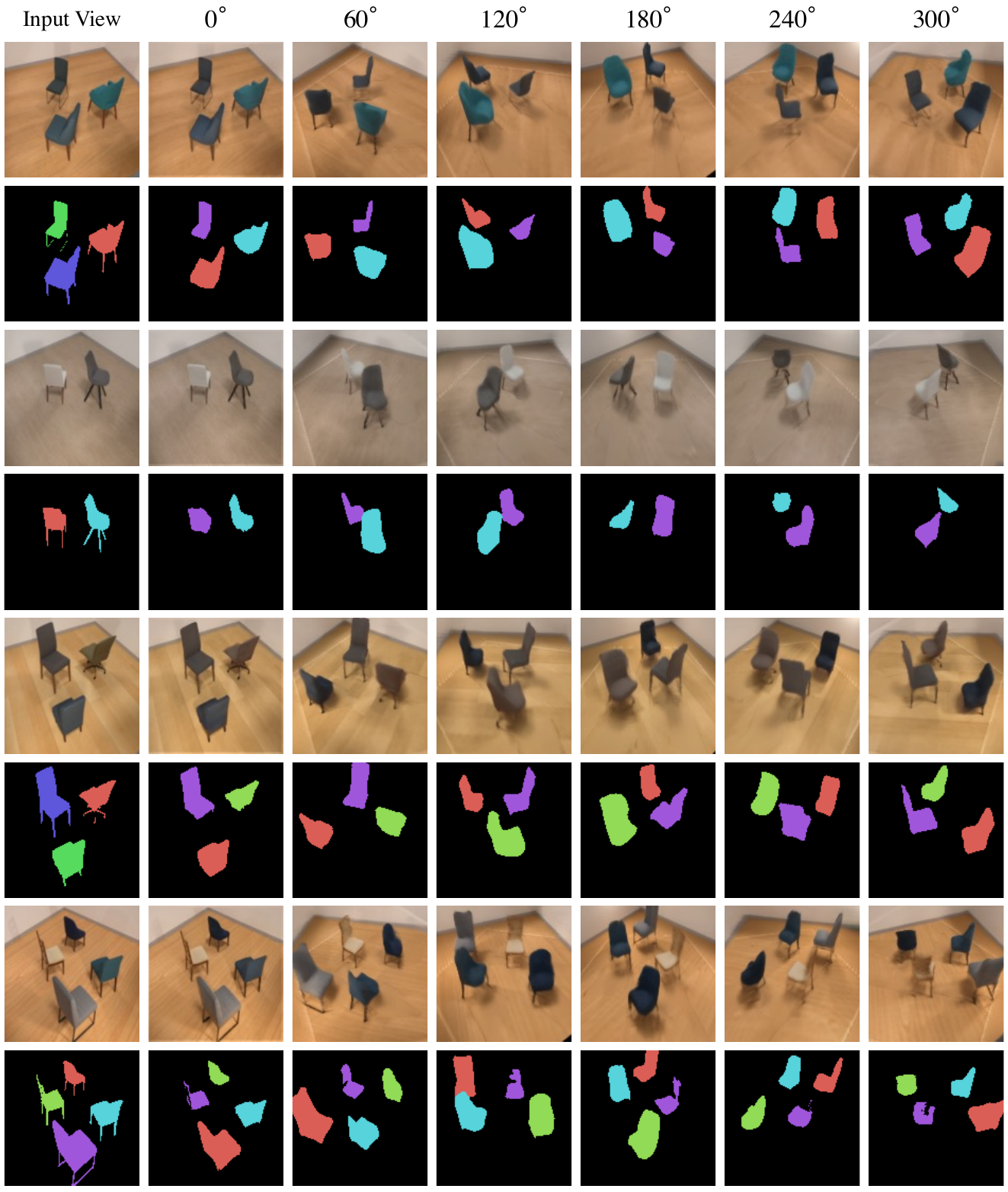}}
    \caption{Novel view synthesis and unsupervised segmentation on Room-Texture.}
    \label{fig:supp_texture}
\end{figure*}

\paragraph{Inference efficiency} While our \model has significantly improved training efficiency compared to other 3D object-centric models, its inference efficiency is not satisfactory compared with light field methods (\eg, COLF and OSRT). The primary reason for this issue is that NeRF representations require the sampling of a large number of points with expensive computations, most of which are wasted on irrelevant vacant points. Although light field methods, such as OSRT, are very efficient for inference, they lack the use of 3D information and require a lot of data and computation commands to overcome the overfitting problem. Some recent works, such as those based on point clouds~\cite{xu2022point}, surfels~\cite{gao2023surfelnerf}, and Gaussian Splatting~\cite{kerbl20233d, zheng2023gps}, have demonstrated high efficiency for inference and good utilization of 3D information, which could be integrated into future work to improve the inference efficiency.
\paragraph{Details of Complex Object} As depicted in~\cref{fig:supp_diverse} and~\cref{fig:supp_texture}, the \model encounters challenges in accurately rendering and segmenting chair legs from different angles, particularly when dealing with real chairs in Room-Texture. A primary issue contributing to this difficulty lies in ray sampling. NeRF-based techniques typically employ ray sampling during the training process to reduce computational load. For example, in our case, we sample 1024 rays from an image with $128\times128=16384$ pixels. Consequently, the majority of rays focus on the background and larger objects, leaving finer details like chair legs with limited attention. While increasing the number of sampled rays could address this issue, it would also escalate the computational demands. The integration of Gaussian Splatting\cite{kerbl20233d} has the potential to assist in balancing computational requirements with rendering quality. Moreover, we have observed that this problem exists in other approaches as well. Nevertheless, it appears to be mitigated in uOCF~\cite{luo2024unsupervised} due to its training with the object prior, which could potentially aid our \model in addressing this particular challenge.
% Moreover, we select nearby views to render the target view on ScanNet following previous generalizable NeRF methods, leading to pseudo-3D representations. 
 
\section{Additional Visualizations}\label{app:vis}
We provide more qualitative results of our \model in the following pages.
\begin{figure*}[t!]
    \centering
    \resizebox{\linewidth}{!}{\includegraphics[width=\linewidth]{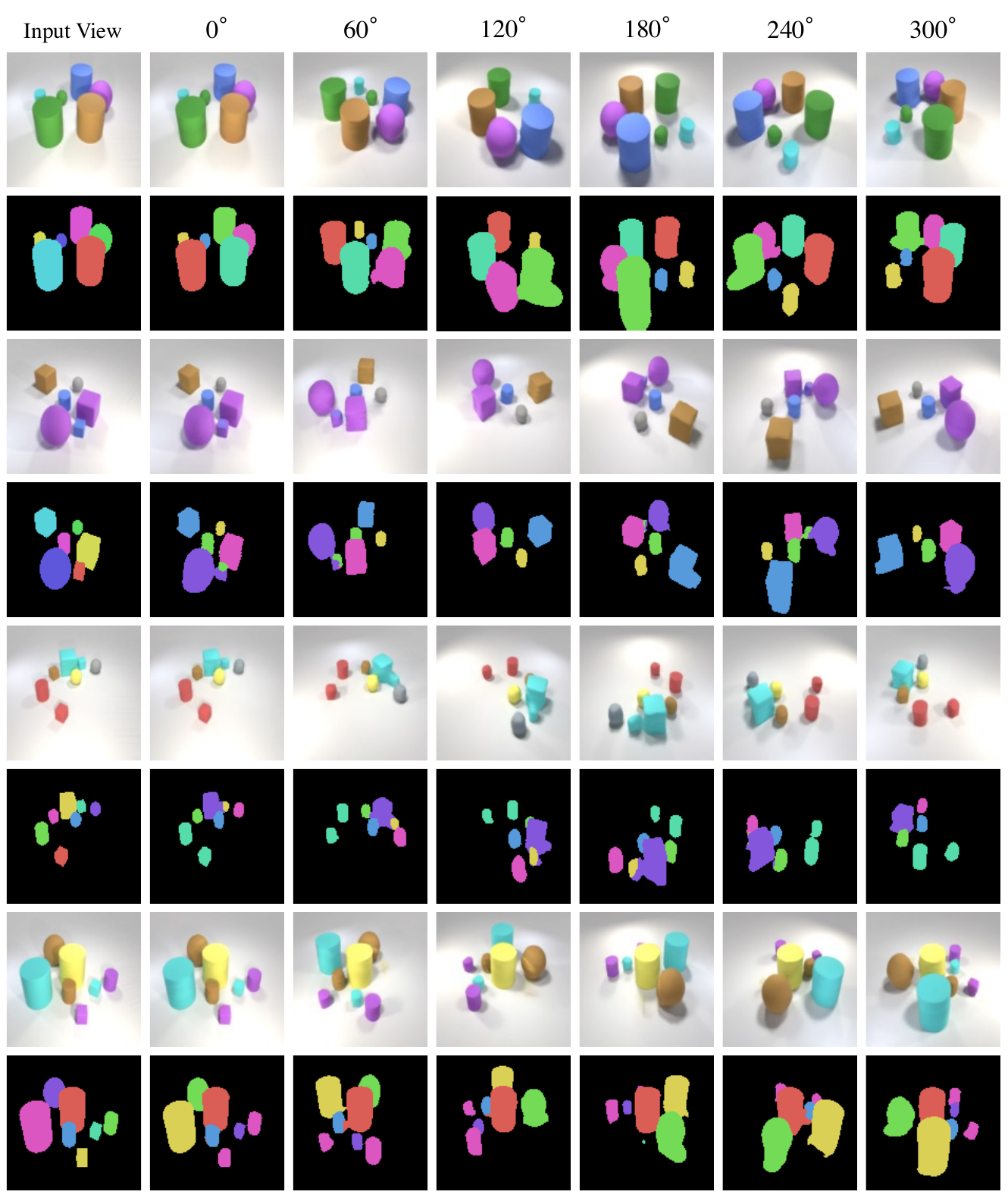}}
    \caption{Novel view synthesis and unsupervised segmentation on CLEVR-567.}
\end{figure*}
\begin{figure*}[t!]
    \centering
    \resizebox{\linewidth}{!}{\includegraphics[width=\linewidth]{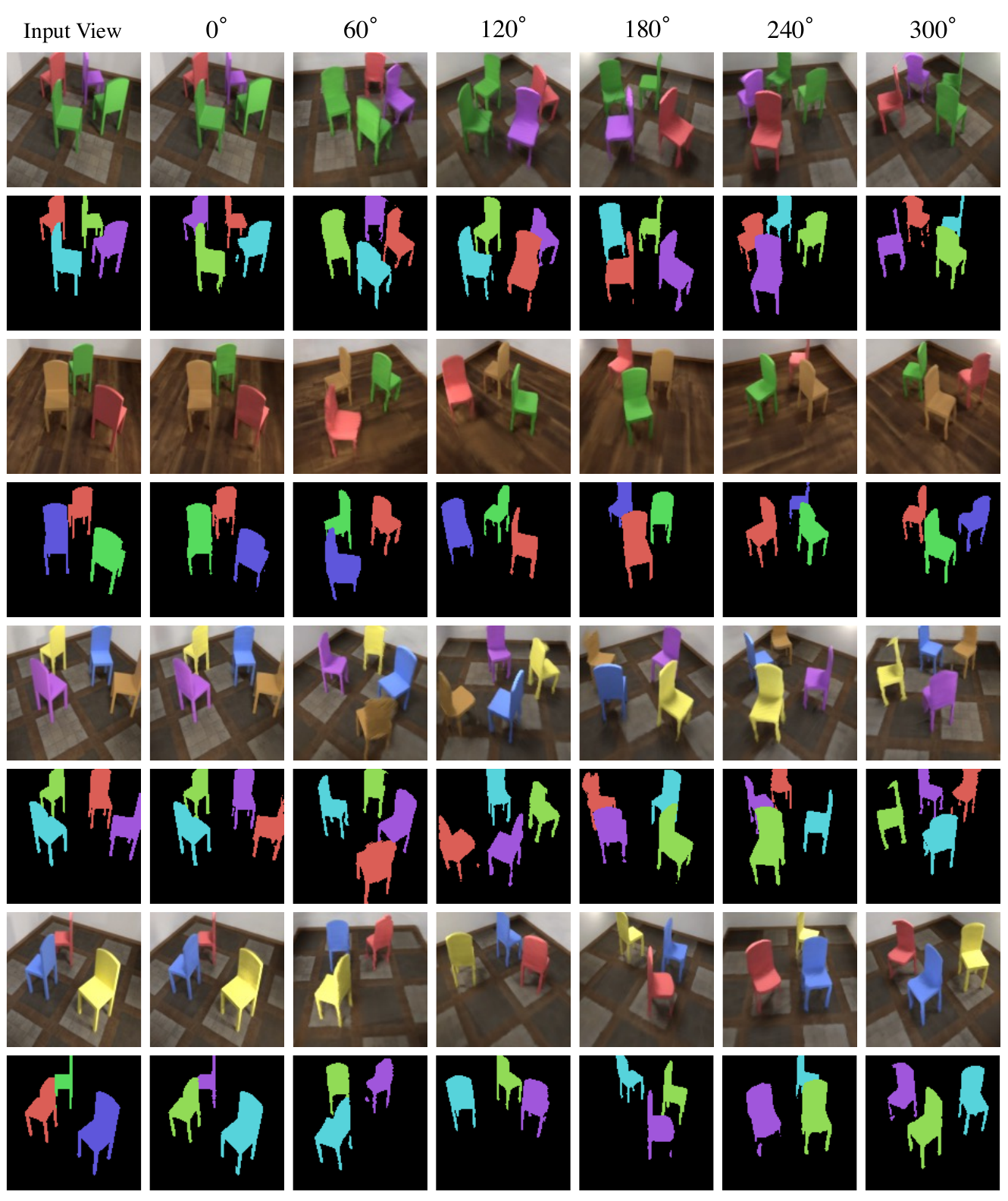}}
    \caption{Novel view synthesis and unsupervised segmentation on Room-Chair.}
\end{figure*}
\begin{figure*}[t!]
    \centering
    \resizebox{\linewidth}{!}{\includegraphics[width=\linewidth]{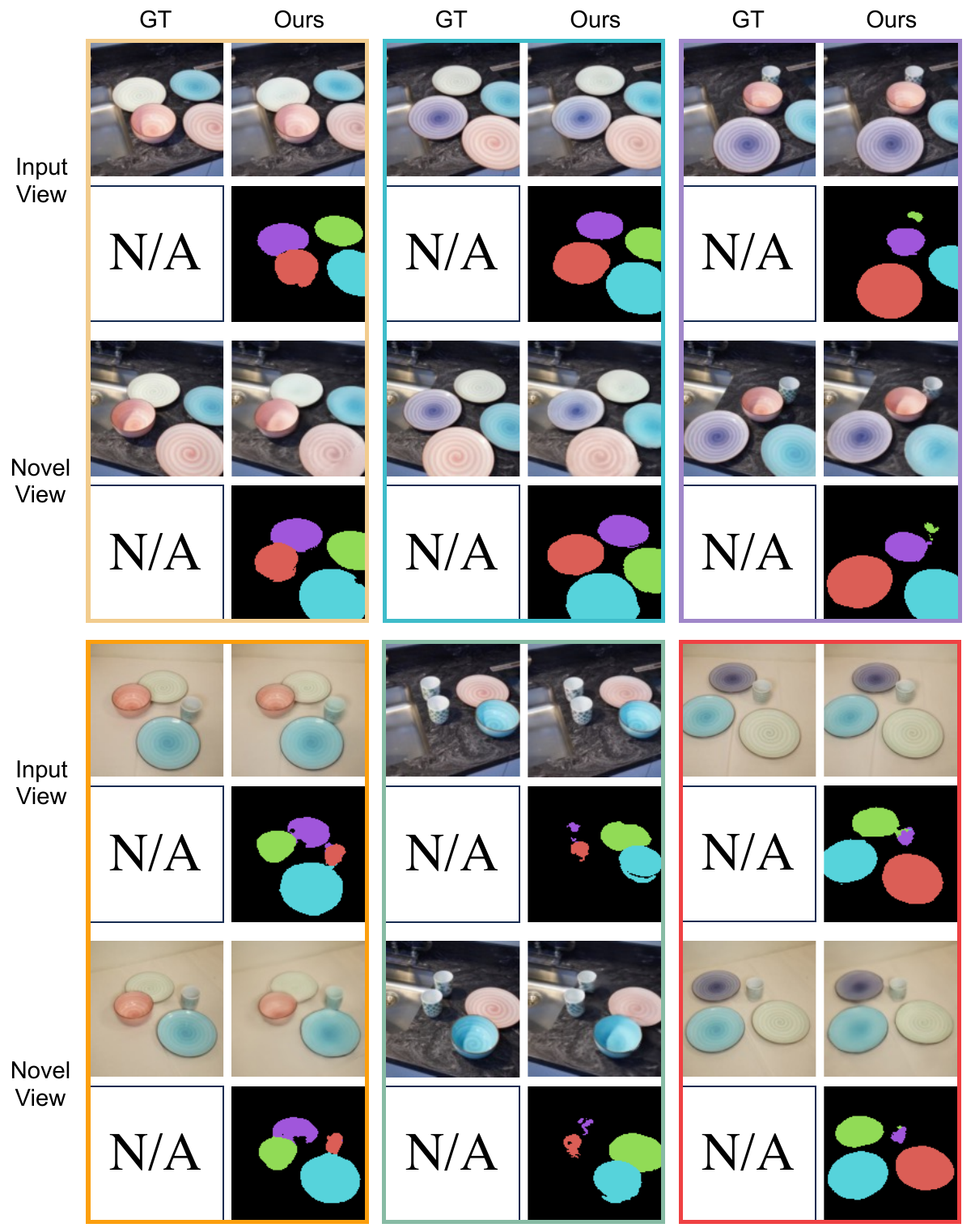}}
    \caption{Novel view synthesis and unsupervised segmentation on Kitchen-Shiny.}
    \label{fig:supp_shiny}
\end{figure*}
\begin{figure*}[t!]
    \centering
    \resizebox{\linewidth}{!}{\includegraphics[width=\linewidth]{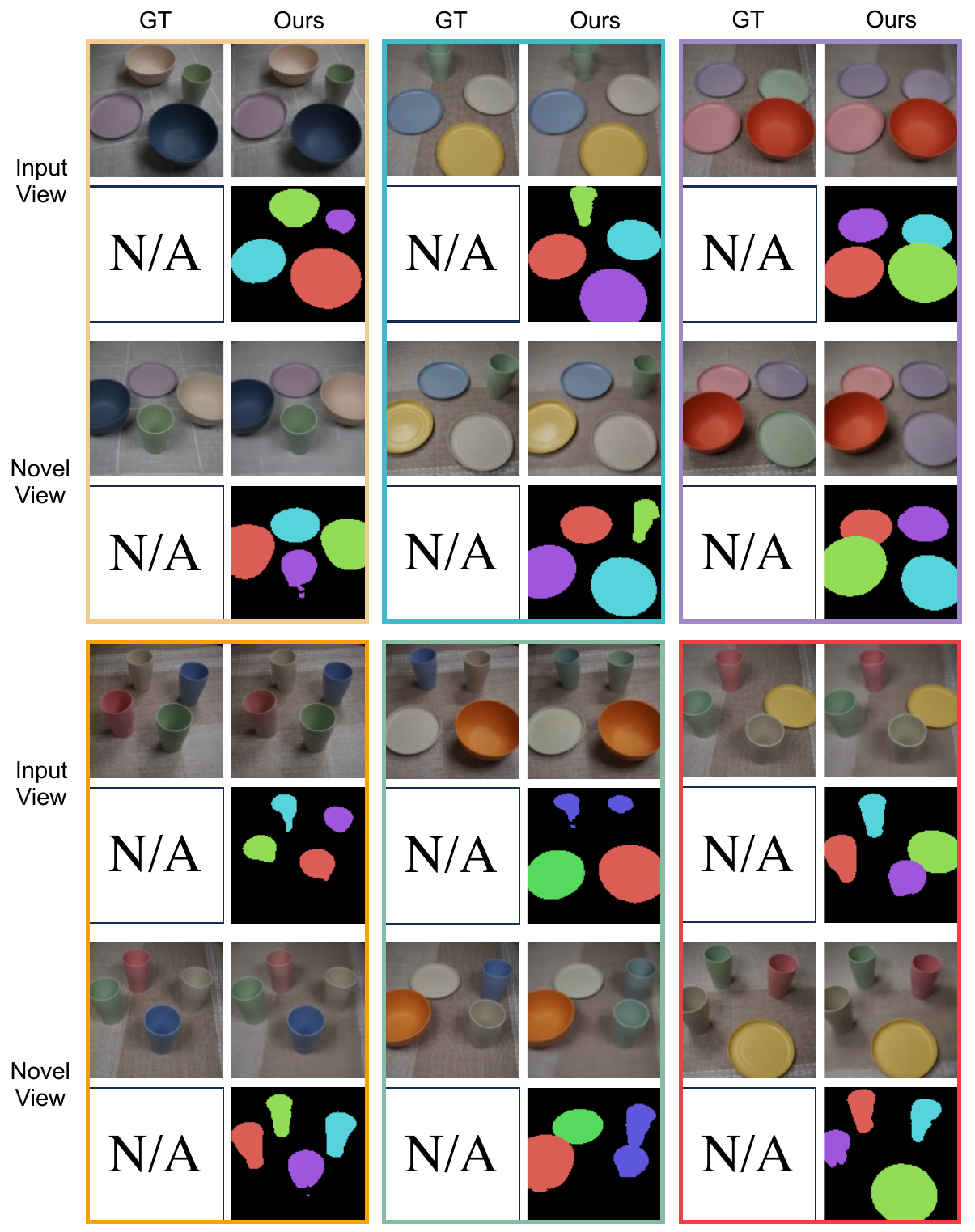}}
    \caption{Novel view synthesis and unsupervised segmentation on Kitchen-Matte.}
    \label{fig:supp_matte}
\end{figure*}
\begin{figure*}[t!]
    \centering
    \resizebox{\linewidth}{!}{\includegraphics[width=\linewidth]{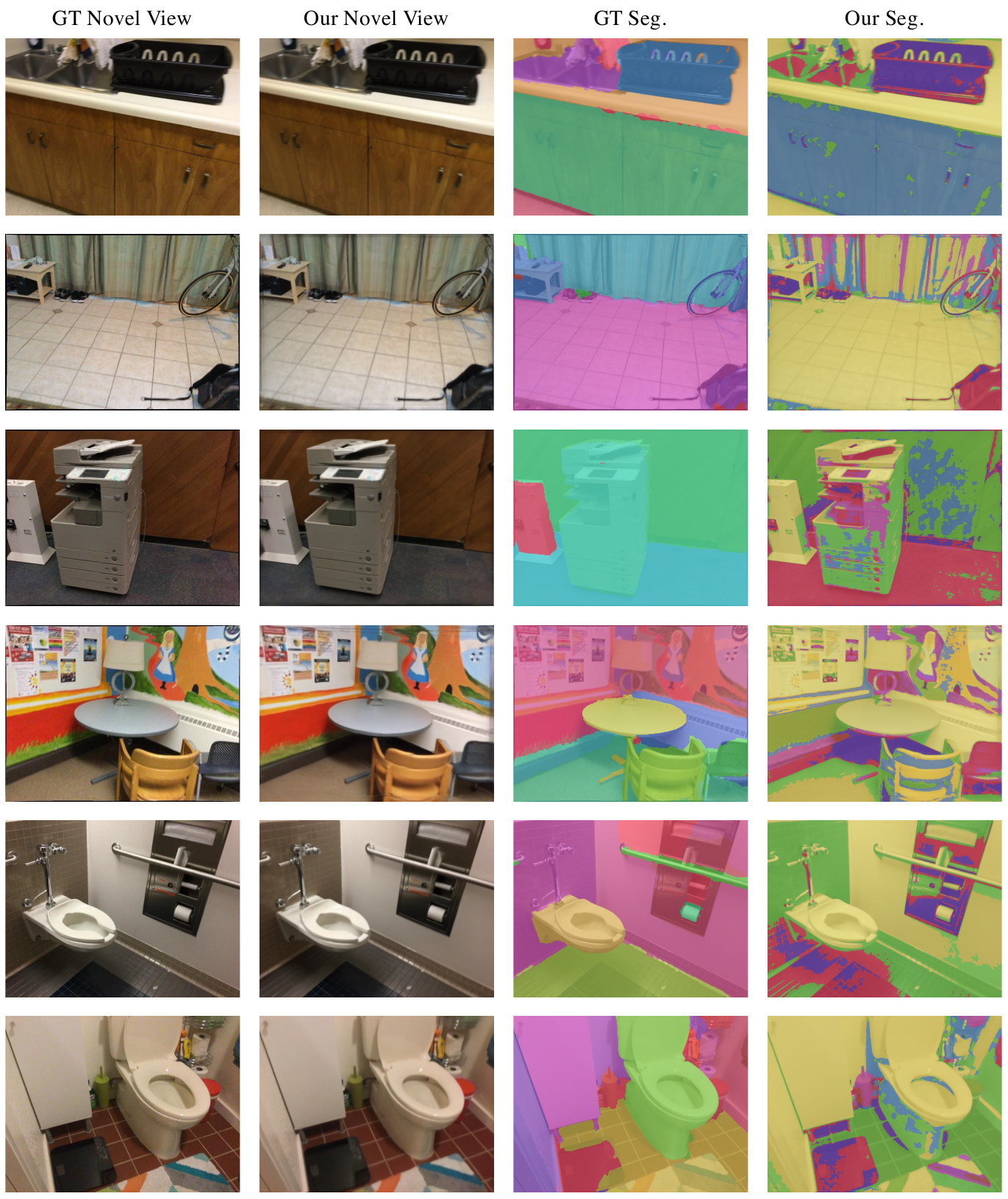}}
    \caption{Novel view synthesis and unsupervised segmentation on ScanNet.}
\end{figure*}
\begin{figure*}[t!]
    \centering
    \resizebox{\linewidth}{!}{\includegraphics[width=\linewidth]{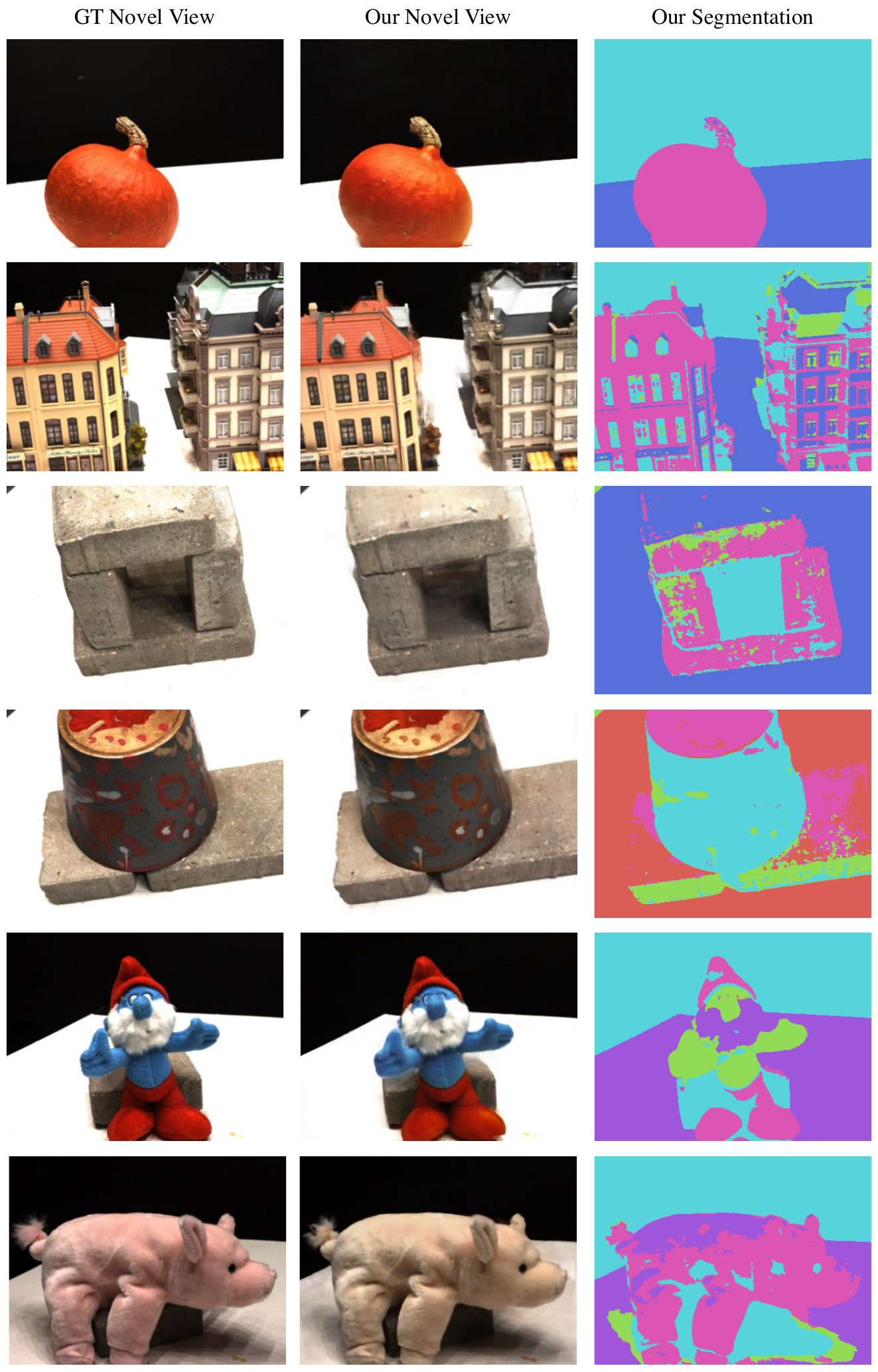}}
    \caption{Novel view synthesis and unsupervised segmentation on DTU MVS.}
\end{figure*}

\end{document}